\documentclass{article}

\usepackage{arxiv}

\usepackage[utf8]{inputenc} 
\usepackage[T1]{fontenc}    
\usepackage{hyperref}       
\usepackage{url}            
\usepackage{booktabs}       
\usepackage{amsfonts}       
\usepackage{amsmath}        
\usepackage{nicefrac}       
\usepackage{microtype}      
\usepackage{cleveref}       
\usepackage{lipsum}         
\usepackage{graphicx}
\usepackage[numbers]{natbib} 
\usepackage{doi}
\usepackage{float}          
\usepackage{subfig}
\usepackage{xcolor}

\title{PruneGCRN: Minimizing and explaining spatio-temporal problems through node pruning}

\newif\ifuniqueAffiliation
\uniqueAffiliationtrue

\ifuniqueAffiliation 
\author{ 
    \href{https://orcid.org/0000-0002-4456-4348}{\includegraphics[scale=0.06]{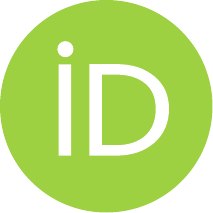}
    \hspace{1mm}Javier García-Sigüenza}\\
    Department of Computer Science and \\
    Artificial Intelligence,\\
    University of Alicante, \\
    San Vicente del Raspeig, Spain.\\
	\texttt{javierg.siguenza@ua.es}\\
	\And
	\href{https://orcid.org/0000-0003-3534-4332}{\includegraphics[scale=0.06]{orcid.pdf}\hspace{1mm}Mirco Nanni}\\
	ISTI-CNR,\\
	Pisa, Italy\\
	\texttt{mirco.nanni@isti.cnr.it}\\
    \AND
    \href{https://orcid.org/0000-0002-2117-0784}{\includegraphics[scale=0.06]{orcid.pdf}
    \hspace{1mm}Faraón Llorens-Largo}\\
    Department of Computer Science and \\
    Artificial Intelligence,\\
    University of Alicante, \\
    San Vicente del Raspeig, Spain.\\
	\texttt{faraon.llorens@ua.es} \\
    \And
    \href{https://orcid.org/0000-0002-2990-1879}{\includegraphics[scale=0.06]{orcid.pdf}
    \hspace{1mm}José F. Vicent}\\
    Department of Computer Science and \\
    Artificial Intelligence,\\
    University of Alicante, \\
    San Vicente del Raspeig, Spain.\\
	\texttt{jvicent@ua.es} \\
}
\else
\fi

\renewcommand{\headeright}{Preprint}
\renewcommand{\undertitle}{}
\renewcommand{\shorttitle}{}

\hypersetup{
pdftitle={PruneGCRN: Minimizing and explaining spatio-temporal problems through node pruning},
pdfsubject={cs.AI},
pdfauthor={Javier García-Sigüenza, Mirco Nanni, Faraón Llorens-Largo, José F. Vicent},
pdfkeywords={Deep learning, Spatio-temporal problems,Explainable artificial intelligence},
}

\begin{document}
\maketitle

\begin{abstract}
This work addresses the challenge of using a deep learning model to prune graphs and the ability of this method to integrate explainability into spatio-temporal problems through a new approach. Instead of applying explainability to the model's behavior, we seek to gain a better understanding of the problem itself. To this end, we propose a novel model that integrates an optimized pruning mechanism capable of removing nodes from the graph during the training process, rather than doing so as a separate procedure. This integration allows the architecture to learn how to minimize prediction error while selecting the most relevant nodes. Thus, during training, the model searches for the most relevant subset of nodes, obtaining the most important elements of the problem, facilitating its analysis. To evaluate the proposed approach, we used several widely used traffic datasets, comparing the accuracy obtained by pruning with the model and with other methods. The experiments demonstrate that our method is capable of retaining a greater amount of information as the graph reduces in size compared to the other methods used. These results highlight the potential of pruning as a tool for developing models capable of simplifying spatio-temporal problems, thereby obtaining their most important elements.
\end{abstract}

\keywords{Deep learning \and Spatio-temporal problems \and Explainable artificial intelligence}

\section{Introduction}

The advancement and spread of AI in recent years has led to increased concern about its transparency and understanding. To address this issue, the field of explainable artificial intelligence (XAI) has become a critical and essential element within the changing landscape of AI development and implementation. Essentially, XAI addresses the “black box” challenge inherent in many AI models by providing users with information about the reasoning behind the decisions made by AI systems \citep{model_interpretability}. 

This increased transparency is crucial for several reasons. On the one hand, it builds trust and facilitates acceptance of AI systems, as it allows users to understand the fundamentals driving the results generated. On the other hand, in high-risk applications such as healthcare, education, finance, and autonomous vehicles, XAI ensures that decisions are not only accurate but also justifiable and free of bias. By revealing potential biases and errors in models, it contributes to the development of fairer and more responsible AI. 

Beyond simply predicting values, XAI also allows for a deeper analysis of the system, identifying the most influential elements within it to understand their impact. For example, in traffic forecasting, it can highlight the most important areas for traffic prediction, allowing the consequences of changes in these areas to be explored. This growing interest in XAI is generating new techniques and examining its potential, consolidating its importance in the AI landscape.

Explainable artificial intelligence addresses two main challenges: interpretability and explainability. Interpretability focuses on designing models with architectures based on logic that is understandable to humans, making them intrinsically interpretable. These models are often referred to as “white box” models. In contrast, explainability focuses on providing insight into the behavior of “black box” models, which are those whose complexity prevents them from being understood by humans. The concept of “gray boxes” has also emerged, describing models that incorporate a certain level of understandability \citep{gray_boxes}. Currently, achieving reliable explainability remains a major obstacle in problems implementing AI, raising several unresolved questions.

The different XAI methods vary depending on the problem to be addressed, as these techniques must be adapted to the models, which in turn must be adapted to the problems. This work seeks to address explainability in spatio-temporal problems, which requires integrating both the spatial and temporal dimensions. These problems are commonly represented by a sequence of graphs. The traffic problem stands out as a particularly relevant example, given its impact on both urban and freight transport. Today, this relevance is amplified by the proliferation of data flows from sources such as road sensors, public transport monitoring, and mobile applications. As a result, there is a large amount of available literature and datasets that support the development and validation of predictive models. 

Among the various approaches to handling spatial and temporal data, models that combine Graph Neural Networks (GNNs) \citep{GNN_Origin} and Recurrent Neural Networks (RNNs) \citep{RNN_Origin} have proven to be effective, standing out from the other options. However, the greater complexity introduced by these models presents a significant challenge when it comes to integrating explainability. Specifically, when a GNN generates a prediction, it is essential to determine the influence of each node in the input graph. Similarly, the addition of temporal data introduces another layer of complexity, as it is necessary to understand how the temporal component affects the prediction. Addressing both issues requires a comprehensive examination of the inner workings of the model.

There are several approaches that aim to provide explainability to deep learning models. The main methods include attention mechanisms, prototype networks, fuzzy systems, and post-hoc explainability methods. Attention mechanisms have gained popularity due to their ability to capture complex relationships between embeddings \citep{Transformer_origin}, while offering explainability by highlighting the most influential feature combinations. However, there are limitations to the explainability of attention mechanisms \citep{is_attention_explanation}, such as the possibility of attention focusing on irrelevant tokens due to “combinatorial shortcuts” \citep{why_not_interpretable} and the difficulty of distinguishing harmful from beneficial trends \citep{negative_attention}. Consequently, determining the components relevant to prediction becomes more complex. Furthermore, these methods were originally designed for other problems, so they need to be adapted to integrate into spatio-temporal problems. 

As for prototype networks \citep{prototype_networks_origin}, they are based on the principle that an embedding represents a prototype for each class, allowing entries to be grouped into clusters derived from these. In the case of prototype networks applied to GNNs, Prototype Graph Neural Network (ProtGNN) \citep{ProtGNN} implements this approach. Explainability is derived from comparing the learned prototypes within a latent space and identifying similar subgraphs within the input graph. Due to this architecture, ProtGNN focuses on classification tasks and does not incorporate the temporal dimension, relying solely on spatial information.

Although models based on fuzzy systems have inherent interpretability properties, their predictive accuracy is often lower, leading to their integration with deep learning models \citep{fuzzy_deep_neural_networks_survey}. The combination of these approaches often results in greater model complexity and higher computational costs, without substantially improving interpretability, making it necessary to find a balance between accuracy and explainability. Furthermore, research focused on spatio-temporal forecasting with fuzzy systems \citep{fuzzy_traffic_2020} prioritizes accuracy over interpretability, which diminishes the advantages of fuzzy system-based models as their accuracy increases. Consequently, the integration of GNNs and RNNs continues to represent the state of the art when it comes to spatio-temporal problems.

When considering post-hoc explainability techniques for GNNs, despite the relevance of the spatio-temporal prediction field and the accuracy they are capable of achieving, integrating such techniques is challenging due to the limited options available. Implementing these techniques often compromises performance in exchange for a certain degree of explainability, and they tend to only take the spatial dimension into account.

For this reason, models such as Self Explainable Graph Convolutional Recurrent Network (SEGCRN) \citep{SEGCRN} have emerged, which seek to adapt post-hoc techniques so that they can be integrated into the architecture of the model itself, thus obtaining gray box models. In this way, the limitations and penalties of post-hoc techniques can be mitigated while obtaining explainability. However, this explainability focuses on discarding the edges of the graph that the model considers superfluous and, based on the analysis of these, understanding the behavior of the model.

Although these alternatives provide different methods to provide explainability, they all pursue the same goal: to understand the behavior of the model through different techniques that allow analyzing how it uses the information it receives. In contrast, the model proposed in this work aims to be able to discard the least relevant nodes in the graph. Thus, it seeks to explore the possibility of changing the type of explainability obtained. By discarding superfluous edges, the transmission of information between the nodes of the model is analyzed, and therefore, the behavior of the model. On the other hand, by discarding nodes, the most superfluous information of the problem is discarded, thus analyzing the problem itself. In this way, the paradigm used changes, seeking to ensure that the explainability of the model helps to understand the problem. To this end, we propose Prune Graph Convolutional Recurrent Network (PruneGCRN), a model capable of maximizing accuracy while discarding nodes during training.

Thus, following the premise that those nodes that provide more information are the most relevant, and that those that provide more information are the ones that allow for better prediction, the aim is to obtain the most important elements of the problem. In this work, we have used traffic data in which each node represents a radar. Therefore, by selecting the most important nodes, we are identifying the most important sensors for monitoring traffic. In this way, PruneGCRN allows us to analyze the most relevant areas for traffic by identifying the most important radars for the problem.

To introduce our work, Section \ref{Preliminaries} presents the problem of spatio-temporal forecasting and explainability. Section \ref{Model_Architecture} describes the model architecture and its components. Section \ref{Model_validation} defines the experiments carried out and the results obtained. Section \ref{Analysis} analyzes the results generated by the model and the explainability obtained. Finally, Section \ref{Conclusions} presents the conclusions obtained.

\section{Preliminaries}\label{Preliminaries}

This paper explores the integration of explainability into a predictive model for spatio-temporal forecasting using graph pruning. To do this, it is first necessary to understand the different models that exist for spatio-temporal forecasting. Next, it is necessary to understand the different existing techniques for providing explainability and, finally, whether these can be adapted to perform graph pruning. Therefore, this section presents the basic concepts of spatio-temporal forecasting, explainability techniques, and existing graph pruning techniques for GNNs. This allows us to introduce the model architecture detailed in the following section.

\subsection{Spatio-temporal forecasting}

Convolutional Neural Networks (CNNs) and RNNs excel at detecting patterns within Euclidean spaces. However, many real-world problems use data represented through non-Euclidean spaces, such as graphs. This change in data representation presents challenges for many deep learning models. Specifically, the variability in the number of neighbors associated with each node in a graph complicates the application of the operations. GNNs were developed to address this problem by developing an architecture tailored to the problem. Among the different types of GNNs that exist, Graph Convolutional Networks (GCNs) \citep{GCN_Origin} are the most popular. These use a set of convolutions to extract the characteristics of the nodes. The combination of graphs with time series creates spatio-temporal problems, relevant to a wide range of applications, including traffic forecasting \citep{gnn_traffic_survey}, fraud detection \citep{gnn_fraud_detection}, social networks \citep{gnn_social_networks}, keyword search \citep{gnn_keyword_search}, and electricity consumption \citep{gnn_electricity}.

The problem of traffic forecasting represents the most studied application of spatio-temporal data, which has led to a considerable set of models and datasets for validating and comparing results. While traditional methods have been used, including statistical regression models such as Autoregressive integrated moving average (ARIMA) \citep{ARIMA_Kalman} and Vector Autoregression (VAR) \citep{VAR}, as well as machine learning methods such as those based on Support Vector Regression (SVR) \citep{SVR}, deep learning models have demonstrated higher accuracy. Specifically, architectures that combine RNNs and GCNs have achieved the best performance by effectively leveraging both temporal and spatial data.

Several models have arisen that combine both architectures. Among these are Diffusion Convolutional Recurrent Neural Network (DCRNN) \citep{DCRNN}, which employ diffusion convolutions together with RNNs to capture both temporal and spatial data through an encoder-decoder structure. Graph WaveNet \citep{GraphWaveNet} uses stacked dilated 1D convolutions combined with GCN layers to process spatial information. Spatio-Temporal Graph Convolutional Networks (STGCN) \citep{STGCN} integrate various blocks incorporating one-dimensional convolutions, gated linear units, and GCN layers to capture temporal and spatial data simultaneously. Attention Based Spatial-Temporal Graph Convolutional Networks (ASTGCN) \citep{ASTGCN_R} employ attention mechanisms alongside convolutions to process spatio-temporal data. Spatial-Temporal Fusion Graph Neural Networks (STFGNN) \citep{STFGNN} learn hidden spatio-temporal dependencies using a parallel operation applied to spatial and temporal graphs, generated using a data-driven method. Spatial-Temporal Graph Ordinary Differential Equation Networks (STGODE) \citep{STGODE} capture dynamic spatio-temporal dependencies using tensor-based ordinary differential equations. Finally, Adaptive Graph Convolutional Recurrent Network (AGCRN) \citep{AGCRN} uses two modules to learn specific characteristics of the nodes and detect spatial and temporal correlations within the time series data. 

Thus, it is clear that there is a wide variety of models for forecasting spatio-temporal problems. However, among those mentioned, AGCRN is the one with the most interesting properties, as it offers great flexibility. That is why its approach of using node embeddings to generate filters and graphs is used in our model. 

\subsection{Explainability techniques}

Explainability in GNNs is an emerging field, currently characterized by a limited set of techniques designed to address the “black box” nature of these models. A key challenge is the lack of standardized metrics to evaluate and compare these methods \citep{explainability_gnns, explainability_custom_metrics}. Despite efforts to develop comparative metrics \citep{graph_fram_ex}, there is still no standard. A commonly used metric, fidelity, quantifies the difference in prediction accuracy between the result of a model with the original graph and its result with a modified graph in which the most influential edges have been removed. It is defined as:

\begin{equation}
\label{eqn:fidelity}
\begin{aligned}
\textit{Fidelity} = \frac{1}{N}\sum_{i=1}^{N} | f(G_{i}) - f(G^{1-m}_{i}) |,\\
\end{aligned}
\end{equation}

where $f(\cdot)$ represents the inference made by the model, $N$ is the number of nodes, $G_i$ is the unmodified graph, and $G^{1-m}_i$ is the graph modified by removing the most important edges.

In contrast, infidelity evaluates the divergence in prediction accuracy between the result of the original graph and the one generated using only the edges considered most important by the explainability technique:

\begin{equation}
\label{eqn:infidelity}
\begin{aligned}
\textit{Infidelity} = \frac{1}{N}\sum_{i=1}^{N} | f(G_{i}) - f(G^{m}_{i}) |,\\
\end{aligned}
\end{equation}

where $f(\cdot)$ denotes the inference of the model, N represents the number of nodes, $G_i$ means the unaltered graph in instance i, and $G^{m}_{i}$ represents the refined graph by excluding the most superfluous edges.

Therefore, the goal is to achieve a high fidelity score, which means that removing critical edges using the explainability method results in a noticeable alteration of the prediction, while a low infidelity value indicates that less significant edges have minimal impact on prediction accuracy \citep{gnn_activation_rules}.

It should be noted that equations \ref{eqn:fidelity} and \ref{eqn:infidelity} differ from those presented in other publications \citep{explainability_2020_gnns} due to the inclusion of the absolute value. When used for prediction rather than classification, it is the difference between values that matters, not their sign.

Another metric used is sparsity, which measures the reduction in the number of edges within the graph. Therefore, a smaller number of edges is preferred when applying explainability methods, as it suggests that the information essential for the prediction of the model is contained in that reduced subgraph. Sparsity is defined as:

\begin{equation}
\label{eqn:sparsity}
\textit{Sparsity} = 1 - \frac{m}{M},
\end{equation}

where $m$ is the number of edges in the subgraph and $M$ is the number of edges in the original graph.

However, for this work, the Fidelity and Infidelity metrics are not useful, as they are designed for post-hoc methods that modify the graph after training the model. In our case, the model selects the nodes during training, so the model is optimized for that subset of nodes, and using the original graph loses its sense, since the model has not been trained for it. Therefore, sparsity has been used instead, along with different accuracy metrics, which do allow to evaluate the impact of the limitations on the number of nodes in the prediction. However, in this case, sparsity measures the difference between nodes rather than edges.

The techniques used to make deep learning models more explainable are usually post-hoc methods, which are applied to models once they have already been trained. These methods are divided into two main groups:

\begin{itemize}
\item \textbf{Methods not designed for GNNs}: These models were originally designed for other types of architectures and therefore need to be adapted to GNNs.
\item \textbf{GNN-Specific methods}: These methods were designed specifically for GNNs and therefore tend to provide better explainability.
\end{itemize}

To evaluate an explainability method, a GNN model is needed to which it can be applied. These are typically evaluated using GCNs, Graph Attention Networks (GAT) \citep{GAT_Origin}, Graph Sample and Aggregate (GraphSAGEs) \citep{GraphSage_Origin}, and Graph Convolutional Network via Initial residual and Identity mapping (GCNII) \citep{GCNII_Origin}.  Applying these methods can be complicated, as it requires us to choose the most appropriate ones and often adapt their implementation to fit the model.  Furthermore, as seen in \citep{traffic_explainability}, the use of post-hoc methods can reduce the accuracy of the model and increase the complexity of the problem.

Therefore, there are different models that have attempted to integrate explainability into the architecture of the model itself. One of these architectures is fuzzy systems, which are based on fuzzy logic. These allow working with degrees of truth instead of strictly binary values. For example, fuzzy systems can formulate rules such as: if variable A is high and variable B is medium, then the output is low. Terms such as high, medium, or low are represented by membership functions with ranges from 0 to 1. This allows for modeling the uncertainty and vagueness inherent in many real-world problems. In addition, fuzzy reasoning follows an interpretable rule structure, close to human language, which makes it a naturally transparent tool.

However, although models based on fuzzy systems have inherent interpretability properties, their predictive accuracy is often lower, which has led to their integration with deep learning models \citep{fuzzy_deep_neural_networks_survey}. The combination of these approaches often results in greater model complexity and higher computational costs, without substantially improving interpretability, making it necessary to find a balance between accuracy and explainability. Furthermore, research focused on spatio-temporal forecasting with fuzzy systems \citep{fuzzy_traffic_2020} prioritizes accuracy over interpretability, diminishing the advantages of fuzzy system-based models as their accuracy increases.

Prototype networks are a type of architecture designed to improve the interpretability of deep learning models, especially in classification tasks. The central idea is that the model learns a set of prototypes, which are latent representations of categories learned during training. When a new sample is introduced, the model projects it into the same latent space and calculates its similarity to existing prototypes. The final decision of the model is based on which prototypes are closest to the sample, so that each prediction can be explained through its similarity to a prototype.

Thus, the usefulness of this model in terms of explainability lies in the fact that the use of prototypes can be analyzed. In this way, instead of relying solely on post-hoc methods, the architecture itself offers intrinsic interpretations: each decision is based on similarities with representative examples that the model has learned throughout training. In the case of prototype networks applied to GNNs, we find Prototype Graph Neural Network (ProtGNN) \citep{ProtGNN}, where explainability is derived from comparing the prototypes learned within a latent space and identifying subgraphs similar to the input graph. A limitation of this architecture is that it is focused on classification and does not apply to forecasting. Furthermore, this model does not incorporate the temporal dimension, relying solely on spatial information.

Attention mechanisms were created to allow deep learning models to focus on the most relevant parts of an input when making decisions, rather than processing all the information uniformly. In terms of explainability, these attention weights are commonly interpreted as a relevance map, showing where or when the model focuses when making a prediction. This allows us to visualize which parts of the data had the most impact on the final decision, offering a first level of transparency. However, it is important to note that attention does not always equate to causal explanation, but rather provides an intuitive and quantitative clue about the model's internal information flow, which is valuable for debugging, validation, and hypothesis generation in complex problems.

However, there are limitations in the explainability of attention mechanisms \citep{is_attention_explanation}, such as the possibility that attention may focus on irrelevant tokens due to “combinatorial shortcuts” \citep{why_not_interpretable} and the difficulty of distinguishing harmful trends from beneficial ones \citep{negative_attention}. As a result, determining the components relevant to prediction becomes more complex. Furthermore, these methods were originally designed for other problems, so they must be adapted to integrate into spatio-temporal problems. It is also important to note that the explainability of attention mechanisms is a secondary property. That is, they are not typically used for explainability, but rather for their ability to predict. Therefore, explainability is not optimized, and no attempt is made to maximize it.

Finally, Self Explainable Graph Convolutional Recurrent Network (SEGCRN) \citep{SEGCRN} is a deep learning model designed specifically to provide explainability to spatio-temporal problems. To do this, it seeks to discard the least relevant edges of the graph, thereby allowing us to understand how it uses the information based on the analysis of these edges. To achieve this, during model training, it optimizes both reducing the number of edges and increasing accuracy. In this way, it integrates explainability into its architecture, training considering this while maintaining its accuracy.

However, although the SEGCRN approach provides explainability for spatio-temporal problems, it focuses on understanding the behavior of the models, not the problem itself. That is why this paper proposes PruneGCRN, which uses the foundations of the previous model but seeks to provide explainability for the problem. To do this, PruneGCRN seeks to minimize the number of nodes in the graph while maximizing prediction accuracy. The aim is to identify the most relevant elements of the problem, on the premise that these will provide the most information and therefore allow for greater accuracy while limiting the number of nodes that the model can use.

\subsection{Graph pruning}

When using GNNs, it is common to encounter limitations when working with large graphs, which is why various attempts have been made to apply graph pruning or alternative methods to GNNs. On the one hand, there are methods that modify the original graph by removing nodes or edges. On the other hand, as an alternative, there are methods that, instead of modifying the graph directly, select neighboring nodes to reduce the variables without modifying the original graph. In both cases, either directly by modifying the original graph or indirectly by selecting a subset without modifying the original graph, the aim is to simplify the problem by reducing the edges or nodes of the graph. Among the models that use these techniques are FastGCN \citep{FastGCN}, GraphSAGE \citep{GraphSage_Origin}, and DyGNN \citep{DyGNN}.

First, FastGCN does probability-based sampling that picks only a fraction of edges in each layer to compute convolutions. This really cuts down on the number of neighbors expanded at each step and, therefore, the computational cost. In addition, GraphSage also follows a sampling strategy to select a subset of edges, but unlike FastGCN, which applies this sampling globally, GraphSage applies it at the node level. Thus, both use sampling strategies that do not modify the original graph. 

In contrast, DyGNN prunes nodes and edges by measuring the similarity between the embeddings of connected nodes in each layer. To do this, it removes nodes that have representations that hardly change and edges whose information is redundant by connecting highly similar nodes. It is important to note that although DyGNN can delete a node, this does not mean that the model does not use it, as other nodes can continue to make use of its information. Therefore, it is only partially removed, unlike PruneGCRN, where the node information is completely deleted and the model cannot make use of it.

Although these models propose different methods for reducing the nodes or edges used by GNNs, they all take a different approach to that of this paper, as they use pruning to reduce computational cost. Thus, the models are not designed to adapt pruning to different sparsities, forcing complete node removal, in order to allow a study of the ability to provide explainability through pruning, unlike PruneGCRN. Furthermore, all these methods focus solely on the spatial dimension, without taking the temporal dimension into account. 

Thus, there is no established baseline in the field of deep learning with which to compare PruneGCRN. Therefore, to validate the model's ability to select the most relevant nodes, its accuracy was compared when using the nodes selected by itself and when using the nodes selected by other algorithms. This allowed us to establish an initial baseline for the problem. In addition, an analysis of node selection has been carried out, as well as the correlation of error and spatial dimension to verify that pruning does not follow spatial criteria alone, but that the model performs a more complex analysis of the problem.

\section{Model architecture}\label{Model_Architecture}

To understand how traffic flows and predict future patterns, it is important to model the complex relationships between spatial and temporal dimensions using spatio-temporal data. Numerous studies have been conducted to create sophisticated GNN models that identify shared patterns between nodes and use predefined graphs. However, PruneGCRN searches for patterns in all nodes, learning during training which ones are most relevant and generating a graph that only contains these nodes. This approach optimizes accuracy while minimizing the amount of information used.

The PruneGCRN model uses two key modules: Node Adaptive Parameter Learning (NAPL) and Pruned Graph Learning (PGL).

\begin{enumerate}
    \item \textbf{Node Adaptive Parameter Learning (NAPL)}: This module learns filters for specific patterns from node embeddings using shared weights and biases.
    \item \textbf{Pruned Graph Learning (PGL)}: This module infers relevant interdependencies between nodes based on the data and generates a simplified graph during training.
\end{enumerate}

The combination of these modules with Gated Recurrent Unit (GRU) \citep{GRU_vs_LSTM} results in the Prune Graph Convolutional Recurrent Network (PruneGCRN) model.

This architecture, PruneGCRN, generates both the filters and the graph as well as the relevance of the connections between nodes during training. Specifically, it uses learned embeddings of the nodes, denoted as $E_N$. The NAPL module uses these embeddings to generate the filters used for feature extraction. Simultaneously, the PGL module creates a graph with the most relevant nodes, rather than relying on a predefined structure.

The prediction process generates both the predictions and the corresponding explainability mask. First, the data is preprocessed using the formula  $z = \frac{x - \mu}{\sigma}$, where $x$ represents the node values, $\mu$ is the mean, and $\sigma$ is the standard deviation. Next, these processed values are fed into the model, which generates node embeddings that are later used to create the learned graph. Finally, these components are combined to perform the spatio-temporal prediction, resulting in the final output and the explainability mask.

Contrary to explainability techniques with a post-hoc approach, in which a model is trained and then explainability is integrated, PruneGCRN generates explainability during training. This integrated approach means that the mask indicating important nodes is inferred and optimized directly from the data used. Consequently, it is essential to understand in detail the two modules used in PruneGCRN to understand how these values are generated.

\subsection{Node Adaptive Parameter Learning (NAPL)}

Let $G(V, E)$ be an undirected graph, where $V$ is the set of $N$ nodes, $E$ is the set of edges, and $A$ is the adjacency matrix of graph $G$. Since the model uses a GCN, which approximates a first-order Chebyshev polynomial expansion, the
graph convolution operation can be defined as:

\begin{equation}
\label{eqn:chebyshev}
Z = (I_{n} + D^{-\frac{1}{2}} A D^{-\frac{1}{2}} ) X \Theta + b,
\end{equation}
with:
\begin{flalign*}
\hspace*{1cm} 
& Z \in \mathbb{R}^{n\,\times\,f}  \rightarrow  \text{Output of the GCN.} &\\
& I_n \in \mathbb{R}^{n\,\times\,n} \rightarrow \text{Identity matrix of size} n \times n. &\\
& D \in \mathbb{R}^{n\,\times\,n} \rightarrow \text{Degree matrix.} &\\
& A \in \mathbb{R}^{n\,\times\,n} \rightarrow \text{Adjacency matrix of $G$} &\\
& X \in \mathbb{R}^{n\,\times\,c} \rightarrow \text{Input of the GCN layer.} &\\
& \Theta \in \mathbb{R}^{c\,\times\,f} \rightarrow \text{Weights of the GCN Layer.} &\\
& b \in \mathbb{R}^{f} \rightarrow \text{Bias of the GCN layer.} &\\
& n \rightarrow \text{Number of nodes.} &\\
& c \rightarrow \text{Number of input variables.} &\\
& f \rightarrow \text{Number of output variables.} &
\end{flalign*}

From the perspective of a node, this convolution operation transforms an input $X^i \in \mathbb{R}^{1 \times c}$ into $Z^i \in \mathbb{R}^{1 \times f}$, where $\Theta$ and $b$ are constants across all nodes. It is important to note that the normalized adjacency matrix $\mathcal{\tilde{L}}$ is represented by the matrix $D^{-\frac{1}{2}} A D^{-\frac{1}{2}}$, which is fundamental to understanding the contributions of the NAPL and PGL modules within PruneGCRN. While in the original implementation of GCN, the convolution performed by $\Theta$ is the same for each node, in PruneGCRN these change between nodes. Therefore, to mitigate the risk of overfitting, in which a large number of parameters ($n \times c \times f$) could lead to excessive complexity, the model learns two smaller matrices, ultimately generating the parameter matrix $\Theta \in \mathbb{R}^{n \times c \times f}$ through their multiplication. So, $\Theta$ is defined as:

\begin{equation}
\label{eqn:NAPL_theta}
\Theta = E_N \cdot W_N,
\end{equation}

where $E_N \in \mathbb{R}^{n \times d}$ represents the node embedding matrix, and $W_N \in \mathbb{R}^{d \times c \times f}$ represents the set of shared weights.  It is essential to note that $d$ is the size of the embedding and is smaller than $n$, remaining independent of the total number of nodes.  The use of these two matrices minimizes the number of parameters that the model must learn. The same principle applies to the bias term $b$.

Applying this principle, the bias term $b$ is calculated as:

\begin{equation}
\label{eqn:NAPL_bias}
b = E_N \cdot b_N, 
\end{equation}

where $b_N$ is a set of shared biases.

Substituting both into Eq. \ref{eqn:chebyshev}, the convolution can be defined as:

\begin{equation}
\label{eqn:NAPL_chebyshev}
Z = (I_{N} + D^{-\frac{1}{2}} A D^{-\frac{1}{2}} ) X E_N W_N + E_N b_N.
\end{equation}

\subsection{Pruned Graph Learning (PGL)}

When GCNs are used to predict spatio-temporal data, it is often necessary to use a predefined adjacency matrix \(A\) to apply convolutions to the graph. Two types of methods are commonly used to calculate these graphs. The first are based on distance, which defines the graph according to the geographical distance between different nodes. The second are defined by a similarity function, which uses the proximity of the nodes by measuring the similarity of the node attributes. However, with these approaches, the predefined graph may not contain complete information about spatial dependency, nor be directly defined for use in prediction tasks, which can lead to considerable bias. Therefore, Pruned Graph Learning (PGL) generates a graph from the node embeddings it learns during training. In addition, this graph is not limited to representing the relationships between nodes. PruneGCRN aims to be able to apply node pruning during training, thus reducing the graph to the desired size while optimizing accuracy for that graph size. This module represents the most important component and novelty of PruneGCRN, as it is what allows it to perform optimized node pruning during the model training itself, which is applied through a mask generated by the model.

\subsubsection{Mask generation}

\begin{figure}[H]
\centering
\includegraphics[width=0.75\linewidth]{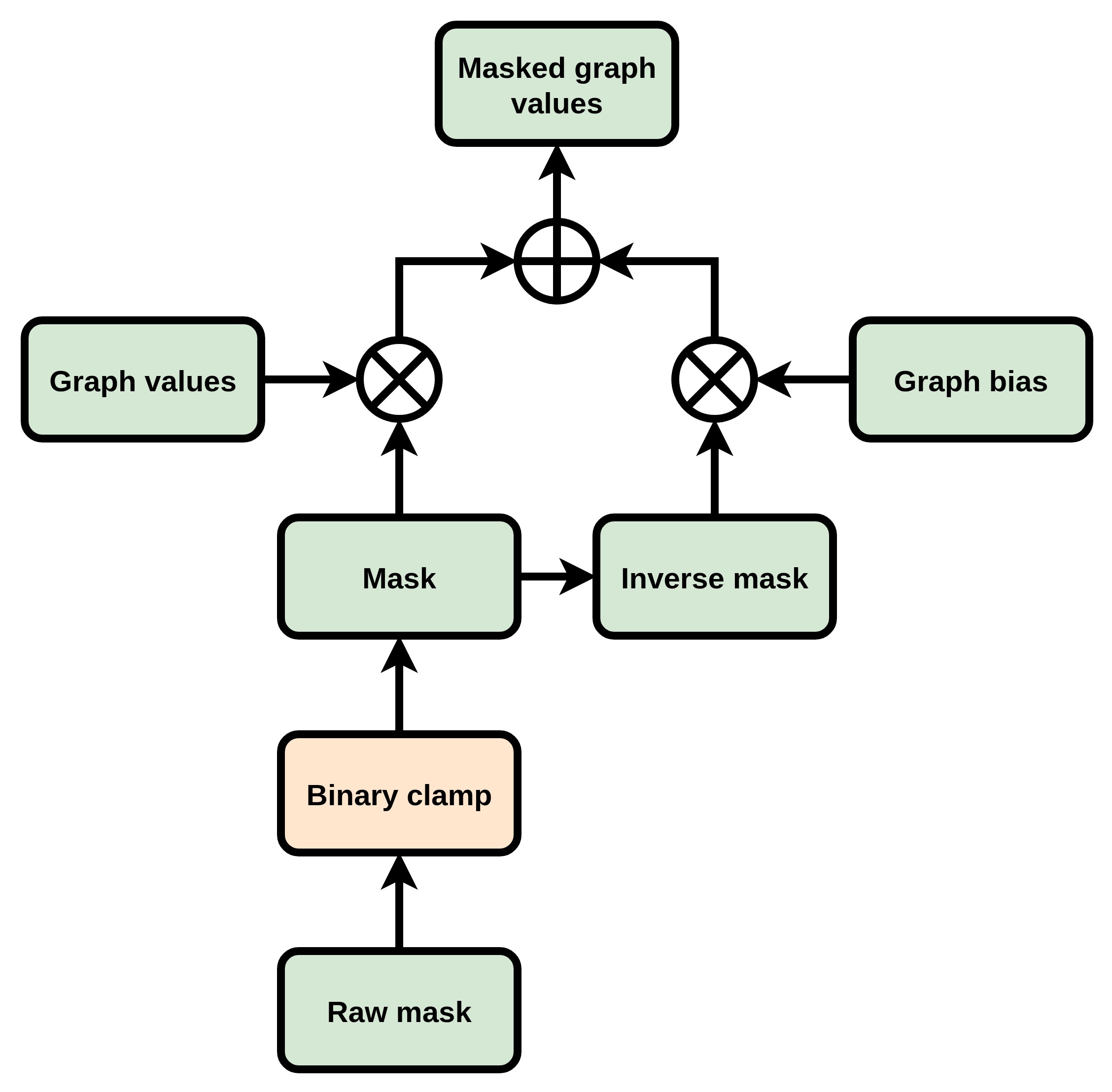} 
\caption{Diagram of the generation of $\tilde{M}$. The parameters are shown in green and the layers in orange.}
\label{fig:mask_generation_flow}
\end{figure}

The mask, denoted as $\tilde{M}$, is generated using the architecture shown in Fig. \ref{fig:mask_generation_flow}. The purpose of this mask is to identify and ignore nodes that have a minimal impact on prediction accuracy. Thus, as they are considered superfluous, they can be discarded. This means that during information propagation, the node value is not taken into account, and instead, a bias learned by the model is used. Therefore, the matrix $\tilde{M}$ composed of the values $\tilde{m}_{i}$ is defined as:

\begin{equation}
\tilde{M} = (\tilde{m}_{1}, \tilde{m}_{2}, ... , \tilde{m}_{n})
\end{equation}
where $\tilde{m}$ is the mask value and $i$ indicates the node.\\

The steps followed to calculate $\tilde{M}$ are shown in Fig. \ref{fig:mask_generation_flow}. First, Raw Mask represents a mask initialized to 1 with float values, so that when training begins, all nodes in the graph are used. Then, Binary Clamp is applied to this mask. This operation consists in setting all values less than 0 to 0, while those greater than 0 become 1. This results in a binary mask, where the values 0 indicate the nodes to be discarded, and the values 1 indicate those to be kept. In turn, the inverse mask is calculated, i.e., the 1s are replaced by 0s and vice versa. This Inverse Mask is used to multiply Graph bias, which is a value learned for each node and is used to replace the masked values of the graph. The purpose of this graph bias is to allow the model to make a more controlled change during training, rather than changing the value of the masked nodes to 0, adding flexibility during training. Finally, the values to be used in the prediction are obtained by combining the values of the graph that have not been masked and the graph bias of the masked nodes.

As previously indicated, in PruneGCRN the mask is calculated using Binary Clamp, which is an alternative to the Hard Concrete technique used by SEGCRN. One of the improvements resulting from the use of Binary Clamp is that PruneGCRN does not exhibit the variability seen in SEGCRN between the training mask and the mask obtained during validation. This is because Hard Concrete uses a normal distribution during training, which is not present during validation, while Binary Clamp always uses Raw Mask values without adding any random components. In this way, the mask is always the same. Moreover, Binary Clamp is simpler to implement and calculate, simplifying its integration.

In addition, this Binary Clamp optimizes node selection in a single step, that is, when performing backpropagation, the model optimizes the Raw Mask values together. In this way, the model does not have an iterative node selection process that greatly increases complexity as the graph size increases.

\subsection{Prune Graph Convolutional Recurrent Network (PruneGCRN)}

By combining all the modules of the model, the convolution is defined as:

\begin{equation}
\label{eqn:Chebyshev_final}
\begin{gathered}
Z = (I_{N} + \mathcal{\tilde{L}}) \tilde{X} E_N W_N + E_N b_N,
\end{gathered}
\end{equation}

where $\tilde{X}$ represents the model input after applying the mask to the nodes and adding the bias. This equation represents the modified GCN architecture with the NAPL and PGL modules. Thus, the addition of NAPL allows specific patterns to be captured in each node and PGL allows a pruned graph to be obtained.

The Eq. \ref{eqn:Chebyshev_final} uses spatial information, but spatio-temporal problems also have a temporal component. Therefore, to add this temporal component and be able to process spatio-temporal information, GRU has been used. To integrate the NAPL and PGL modules into GRU, the fully connected layers used in GRU are replaced by NAPL and PGL.

Finally, the GRU layer with the addition of bias is:

\begin{equation}
\label{eqn:GRU_Origin}
\begin{gathered}
z_t = \sigma (\mathcal{\tilde{L}} [\tilde{X}_{:,t} , h_{t-1}] E_N W_{zr} + E b_{zr} ),\\
r_t = \sigma (I_n [\tilde{X}_{:,t} , h_{t-1}] E_N W_{zr} + E b_{zr} ),\\
\hat{h}_t = \tanh ( [I_n  +\mathcal{\tilde{L}}] [\tilde{X}_{:,t} , r \odot h_{t-1}] E_N W_{\hat{h}} + E_N b_{\hat{h}} ),\\
h_t = z_t \odot \hat{h}_{t-1} + (1 - z_t) \odot \hat{h}_{t-1},\\
\end{gathered}
\end{equation}

where $E_N$ is the node embedding shared by NAPL and PGL, $z_t$ is the update gate, $r_t$ is the reset gate, $h_t$ is the output at step \(t\), $x_t$ is the input at step \(t\), $\hat {h}_t$ is the candidate activation at step \(t\), $W, U, b$ are parameters learned by the model during training, $\odot$ is the Hadamard product, $\sigma$ is the sigmoid function, and ${[\cdot]}$ is the concatenation operator.

To obtain the final model, which will perform the prediction, several layers are concatenated, followed by a final layer that will perform a linear transformation, resulting in a matrix of dimension \( R^{n \times \tau} \), where \( \tau \) is the window size, which indicates the number of steps to be predicted.

\section{Model validation}\label{Model_validation}

This section shows the experiments carried out to validate the model. To this end, the datasets and metrics used are described, and the different results obtained by the model are shown.

\subsection{Datasets}

The PeMSD3, PeMSD4, PeMSD7, and PeMSD8 datasets are the most commonly used datasets when validating traffic forecasting models. These datasets come from the freeway Performance Measurement System (PeMS) \citep{PEMS}, managed by the California Department of Transportation. Each dataset contains measurements from a network of sensors, providing a single variable representing the volume of traffic measured by each sensor, recorded at 5-minute intervals. Datasets vary in terms of the number of sensors and the length of the time series covered.

\begin{itemize}
    \item PeMSD3 includes 358 sensors and covers the period from September 9, 2018, to November 30, 2018.
    \item PeMSD4 includes 307 sensors and covers the period from January 1, 2018, to February 28, 2018.
    \item PeMSD7 includes 883 sensors and covers the period from May 1, 2017, to August 31, 2018.
    \item PeMSD8 includes 170 sensors and covers the period from July 1, 2018, to August 31, 2018.
\end{itemize}

In addition, the PeMS-Bay dataset is used, as it includes the geolocation of the sensors. This feature allows the results to be effectively visualized on a map, facilitating analysis. The PeMS-Bay dataset, obtained from PeMS between January 1, 2017 and May 31, 2017, contains measurements of the average speed of vehicles taken by 325 sensors, recorded at 5-minute intervals.

\subsection{Precision metrics}

To train and validate the model, the Mean Absolute Error (MAE) was selected. This metric is widely used to evaluate the accuracy of predictions, especially in traffic time series, where extreme values are rare. Minimum and maximum speeds, as well as vehicle volumes, tend to fluctuate within predictable ranges, making MAE a suitable choice. The definition of MAE is:

\begin{equation}
\begin{aligned}
\text{MAE} = \frac{1}{N}\sum_{i=1}^{N} |y_i - \hat{y_i}|, \\
\end{aligned}
\end{equation}

where $N$ is the amount of data, and $y_i$ represents the target data, while $\hat{y_i}$ denotes the predicted values. 

To train the model, it was necessary to define a loss function that took into account both the accuracy of the model and the nodes used to make the prediction. To do this, it is first necessary to calculate the mask loss, defined as:

\begin{equation}
\begin{aligned}
M_{\text{Loss}}= \left( \frac{1}{N} \sum_{i=1}^{N} \tilde{M}_{i} \right) - \gamma, \\
\end{aligned}
\end{equation}

where $M_{\text{Loss}}$ represents the mask loss, $\tilde{M} \in \mathbb{R}^{n \times n}$ is the mask generated by PGL, $n$ represents the number of nodes in the graph, and $\gamma$ is a hyperparameter that defines the percentage of nodes that can be used without penalty. Using binary values, the mask loss $M_{\text{Loss}}$ penalizes the number of nodes used, represented by a value of 1, with the aim of maximizing the number of 0 values in the mask.

Finally, based on $M_\text{Loss}$, we can define the loss function used to train the model, which is defined as:

\begin{equation}
\begin{aligned}
\text{Loss} = (\frac{1}{N}\sum_{i=1}^{N} |y_i - \hat{y_i}|) (1 + M_{\text{Loss}}), \\
\end{aligned}
\end{equation}

Therefore, the impact of $M_\text{Loss}$ is conditioned by the change in the prediction error it generates, since a reduction in the use of nodes can lead to greater error depending on its effect on accuracy.

In addition, to validate the model, apart from the MAE, the Root Mean Square Error (RMSE) and the Mean Absolute Percentage Error (MAPE) have also been used, which are defined as:

\begin{equation}
\begin{aligned}
\text{RMSE} = \sqrt{ \frac{1}{N}\sum_{i=1}^{N} {(y_i - \hat{y_i})}^2 }, \\
\end{aligned}
\end{equation}

\begin{equation}
\begin{aligned}
\text{MAPE} =  \frac{100\%}{N}\sum_{i=1}^{N} |\frac{y_i - \hat{y_i}}{y_i}|, \\
\end{aligned}
\end{equation}

\subsection{Experimental settings}

For the experiments, PeMSD3, PeMSD4, PeMSD7, PeMSD7, PeMSD7, and PeMS-Bay datasets were divided into training, validation, and test sets, using a ratio of 6:2:2 respectively, while maintaining the temporal order of the data. The implementation was carried out in Python using PyTorch, and training was performed on an NVIDIA RTX 3090 Ti GPU. RAdam optimizer \citep{RADAM} was used to train the models, and the hyperparameters, as defined in Table \ref{tab:hyperparameters}, were set for each dataset.  The lowest validation error was used as the model selection criterion, and the final performance of the model was evaluated on the test set.

\begin{table}[h]
\centering
\captionsetup{skip=0.4em}
\begin{tabular}{|l|l|l|l|l|l|}
\hline
\textbf{Hyperparameter} & \textbf{PeMSD3} & \textbf{PeMSD4} & \textbf{PeMSD7} & \textbf{PeMDS8} & \textbf{PeMS-Bay} \\
\hline
Batch size & 32 & 32 & 32 & 32 & 32 \\

Embedding dimensions & 10 & 20 & 20 & 20 & 20 \\

RNN Units & 64 & 64 & 64 & 64 & 64 \\

Window & 12 & 12 & 12 & 12 & 12 \\

Horizon & 12 & 12 & 12 & 12 & 12\\

Learning rate & 0.001 & 0.001 & 0.001 & 0.001 & 0.001\\

Early stop & 25 & 25 & 25 & 25 & 25\\

\hline
\end{tabular}

\caption{Hyperparameters used to train PruneGCRN on the different datasets.}
\label{tab:hyperparameters}
\end{table} 

\subsection{Results}

The PruneGCRN approach is novel, as there are no other deep learning models for spatio-temporal problems whose design focuses on finding the best subgraph while optimizing accuracy. Therefore, to validate the capacity of the model to optimize the prediction while using the learned mask, its performance was evaluated by comparing when the model uses the mask learned by itself and when it is provided with a mask generated by two other methods: \textit{random} and \textit{correlation}. The random method corresponds to the generation of random masks and serves to establish a baseline that shows the ability to at least surpass the most basic possible strategy. The correlation method used, inspired by Correlation-based Feature Selection (CFS) \citep{CFS_journal}, follows the premise that those nodes that have the lowest correlation with respect to the others are the most independent and, therefore, may present information that is more difficult to infer based on the other nodes. Thus, for the correlation method, a subset of nodes with the lowest correlation with respect to each other has been chosen, based on the average of the correlations with each of the other nodes in the graph, and these have been used as a mask. When using the \textit{random} or \textit{correlation} methods, the mask is set at the beginning of training and cannot be modified during it.

\setlength{\tabcolsep}{0.2em}
\begin{table}[H]
\centering
\captionsetup{skip=0.4em}
\def\arraystretch{1.3}%
\fontsize{7.8pt}{7.8pt}\selectfont
\begin{tabular}{|c|c|c|cccccc|}
\hline
Pruning & Dataset & Metrics & 0\% & 25\% & 50\% & 75\% & 90\% & 95\%\\
\hline
        &          & MAE & 15.41 $\pm$ 0.09 & 16.67 $\pm$ 0.11 & 17.51 $\pm$ 0.13 & 18.61 $\pm$ 0.16 & 19.72 $\pm$ 0.36 & 20.74 $\pm$ 0.19\\
        & PeMSD3   & RMSE & 27.02 $\pm$ 0.21 & 29.30 $\pm$ 0.26 & 30.80 $\pm$ 0.40 & 32.83 $\pm$ 0.18 & 34.47 $\pm$ 0.62 & 35.93 $\pm$ 0.43\\
        &          & MAPE & 14.87\% $\pm$ 0.21\% & 16.21\% $\pm$ 0.36\% & 16.84\% $\pm$ 0.20\% & 17.66\% $\pm$ 0.21\% & 18.75\% $\pm$ 0.44\% & 19.54\% $\pm$ 0.30\%\\
\cline{2-9}
        &          & MAE & 19.43 $\pm$ 0.10 & 20.60 $\pm$ 0.19 & 21.71 $\pm$ 0.21 & 22.93 $\pm$ 0.25 & 24.01 $\pm$ 0.45 & 24.72 $\pm$ 0.41\\
        & PeMSD4   & RMSE & 32.09 $\pm$ 0.25 & 34.49 $\pm$ 0.46 & 36.52 $\pm$ 0.64 & 38.46 $\pm$ 0.58 & 40.06 $\pm$ 0.82 & 40.87 $\pm$ 0.44\\
        &          & MAPE & 13.06 $\pm$ 0.17\% & 13.90\% $\pm$ 0.31\% & 14.65\% $\pm$ 0.16\% & 15.30\% $\pm$ 0.20\% & 16.00\% $\pm$ 0.32\% & 16.40\% $\pm$ 0.40\%\\
\cline{2-9}
        &          & MAE & 20.74 $\pm$ 0.05 & 23.42 $\pm$ 0.29 & 26.10 $\pm$ 0.36 & 28.93 $\pm$ 0.35 & 30.86 $\pm$ 0.48 & 31.71 $\pm$ 0.43\\
Random  & PeMSD7   & RMSE & 34.74 $\pm$ 0.14 & 41.68 $\pm$ 1.38 & 48.01 $\pm$ 1.54 & 53.84 $\pm$ 1.26 & 57.06 $\pm$ 0.69 & 58.46 $\pm$ 0.38\\
        &          & MAPE & 8.82\% $\pm$ 0.07\% & 10.02\% $\pm$ 0.17\% & 11.16\% $\pm$ 0.19\% & 12.35 $\pm$ 0.17\% & 13.25\% $\pm$ 0.22\% & 13.69\% $\pm$ 0.17\% \\
\cline{2-9}
        &          & MAE & 16.00 $\pm$ 0.11 & 17.93 $\pm$ 0.32 & 19.72 $\pm$ 0.43 & 21.92 $\pm$ 0.64 & 23.77 $\pm$ 1.18 & 25.83 $\pm$ 2.18 \\
        & PeMSD8   & RMSE & 25.46 $\pm$ 0.17 & 29.56 $\pm$ 0.89 & 33.39 $\pm$ 1.17 & 37.72 $\pm$ 1.52 & 40.42 $\pm$ 1.86 & 43.34 $\pm$ 2.70\\
        &          & MAPE & 10.33\% $\pm$ 0.08\% & 11.48\% $\pm$ 0.25\% & 12.38\% $\pm$ 0.17\% & 13.50\% $\pm$ 0.26\% & 14.55\% $\pm$ 0.73\% & 15.56 $\pm$ 1.16\%\\
\cline{2-9}
        &          & MAE & 1.66 $\pm$ 0.01 & 1.85 $\pm$ 0.02 & 2.06 $\pm$ 2.28 & 2.28 $\pm$ 0.02 & 2.48 $\pm$ 0.04 & 2.71 $\pm$ 0.11\\
        & PeMS-Bay & RMSE & 3.82 $\pm$ 0.02 & 4.15 $\pm$ 0.05 & 4.50 $\pm$ 0.06 & 4.91 $\pm$ 0.06 & 5.26 $\pm$ 0.07 & 5.69 $\pm$ 0.21\\
        &          & MAPE & 3.79\% $\pm$ 0.02\% & 4.21\% $\pm$ 0.05\% & 4.73\% $\pm$ 0.08\% & 5.30\% $\pm$ 0.08\% & 5.87\% $\pm$ 0.10\% & 6.45\% $\pm$ 0.26\%\\
\hline
        &          & MAE & 15.41 $\pm$ 0.09 & 16.40 $\pm$ 0.09 & 17.66 $\pm$ 0.11 & 19.69 $\pm$ 0.23 & 21.19 $\pm$ 0.21 & 22.69 $\pm$ 0.31\\
        & PeMSD3   & RMSE & 27.02 $\pm$ 0.21 & \textbf{28.56} $\pm$ \textbf{0.12} & 30.84 $\pm$ 0.31 & 33.90 $\pm$ 0.33 & 36.28 $\pm$ 0.33 & 37.86 $\pm$ 0.43\\
        &          & MAPE & 14.87\% $\pm$ 0.21\% & 15.84\% $\pm$ 0.35\% & 16.90\% $\pm$ 0.29\% & 18.17\% $\pm$ 0.27\% & 19.48\% $\pm$ 0.44\% & 21.56\% $\pm$ 0.60\% \\
\cline{2-9}
        &          & MAE & 19.43 $\pm$ 0.10 & \textbf{20.06} $\pm$ \textbf{0.06} & 21.19 $\pm$ 0.09 & 23.49 $\pm$ 0.28 & 25.95 $\pm$ 0.17 & 27.05 $\pm$ 0.15\\
        & PeMSD4   & RMSE & 32.09 $\pm$ 0.25 & \textbf{32.69} $\pm$ \textbf{0.21} & \textbf{34.35} $\pm$ \textbf{0.28} & 38.12 $\pm$ 0.42 & 42.83 $\pm$ 0.34 & 44.64 $\pm$ 0.31\\
        &          & MAPE & 13.06 $\pm$ 0.17\% & \textbf{13.52\%} $\pm$ \textbf{0.12\%} & 14.32\% $\pm$ 0.17\% & 15.63\% $\pm$ 0.32\% & 17.78\% $\pm$ 0.27\% & 18.50\% $\pm$ 0.16\%\\
\cline{2-9}
        &          & MAE & 20.74 $\pm$ 0.05 & \textbf{21.83} $\pm$ \textbf{0.07} & 25.58 $\pm$ 0.44 & 29.71 $\pm$ 0.39 & 31.35 $\pm$ 0.41 & 36.21 $\pm$ 1.09\\
Correlation & PeMSD7 & RMSE & 34.74 $\pm$ 0.14 & \textbf{36.14} $\pm$ \textbf{0.16} & \textbf{41.07} $\pm$ \textbf{0.74} & 46.65 $\pm$ 0.48 & 53.90 $\pm$ 0.47 & 58.65 $\pm$ 1.10 \\
        &          & MAPE & 8.82\% $\pm$ 0.07\% & \textbf{9.28\%} $\pm$ \textbf{0.04\%} & 10.92\% $\pm$ 0.17\% & 12.59\% $\pm$ 0.19\% & 13.75\% $\pm$ 0.29\% & 17.71\% $\pm$ 1.32\% \\
\cline{2-9}
        &          & MAE & 16.00 $\pm$ 0.11 & \textbf{16.43} $\pm$ \textbf{0.12} & 18.23 $\pm$ 0.12 & 20.64 $\pm$ 0.20 & 23.62 $\pm$ 0.21 & 25.42 $\pm$ 0.64 \\
        & PeMSD8   & RMSE & 25.46 $\pm$ 0.17 & \textbf{26.01} $\pm$ \textbf{0.18} & 28.76 $\pm$ 0.24 & 32.78 $\pm$ 0.38 & 38.96 $\pm$ 0.69 & 40.96 $\pm$ 1.22 \\
        &          & MAPE & 10.33\% $\pm$ 0.08\% & \textbf{10.61\%} $\pm$ \textbf{0.14\%} & 11.38\% $\pm$ 0.10\% & 12.83\% $\pm$ 0.26\% & 13.94\% $\pm$ 0.19\% & 15.16\% $\pm$ 0.58\% \\
\cline{2-9}
        &          & MAE & 1.66 $\pm$ 0.01 & 1.85 $\pm$ 0.01 & 2.10 $\pm$ 0.02 & 3.02 $\pm$ 0.11 & 3.34 $\pm$ 0.11 & 3.69 $\pm$ 0.16\\
        & PeMS-Bay & RMSE & 3.82 $\pm$ 0.02 & \textbf{4.10} $\pm$ \textbf{0.02} & 4.55 $\pm$ 0.03 & 6.26 $\pm$ 0.22 & 6.86 $\pm$ 0.22 & 7.58 $\pm$ 0.25\\
        &          & MAPE & 3.79\% $\pm$ 0.02\% & 4.18\% $\pm$ 0.03\% & 4.79\% $\pm$ 0.02\% & 6.63\% $\pm$ 0.13\% & 7.58\% $\pm$ 0.24\% & 8.67\% $\pm$ 0.16\%\\
\hline
        &          & MAE & 15.41 $\pm$ 0.09 & \textbf{16.26} $\pm$ \textbf{0.16} & \textbf{16.91} $\pm$ \textbf{0.07} & \textbf{17.81} $\pm$ \textbf{0.10} & \textbf{18.79} $\pm$ \textbf{0.13} & \textbf{19.55} $\pm$ \textbf{0.31} \\
        &  PeMSD3  & RMSE & 27.02 $\pm$ 0.21 & 28.67 $\pm$ 0.35 & \textbf{29.69} $\pm$ \textbf{0.19} & \textbf{31.44} $\pm$ \textbf{0.23} & \textbf{33.02} $\pm$ \textbf{0.21} & \textbf{34.01} $\pm$ \textbf{0.37}\\
        &          & MAPE & 14.87\% $\pm$ 0.21\% & \textbf{15.59\%} $\pm$ \textbf{0.24\%} & \textbf{16.36\%} $\pm$ \textbf{0.23\%} & \textbf{17.37\%} $\pm$ \textbf{0.35\%} & \textbf{17.89\%} $\pm$ \textbf{0.31\%} & \textbf{18.36\%} $\pm$ \textbf{0.36\%} \\
\cline{2-9}
        &          & MAE & 19.43 $\pm$ 0.10 & 20.17 $\pm$ 0.10 & \textbf{21.01} $\pm$ \textbf{0.11} & \textbf{21.88} $\pm$ \textbf{0.14} & \textbf{23.06} $\pm$ \textbf{0.18} & \textbf{23.74} $\pm$ \textbf{0.43} \\
        &  PeMSD4  & RMSE & 32.09 $\pm$ 0.25 & 34.17 $\pm$ 0.38 & 35.76 $\pm$ 0.26 & \textbf{37.05} $\pm$ \textbf{0.23} & \textbf{38.72} $\pm$ \textbf{0.25} & \textbf{39.59} $\pm$ \textbf{0.49} \\
        &    & MAPE & 13.06\% $\pm$ 0.17\% & 13.55\% $\pm$ 0.13\% & \textbf{14.13\%} $\pm$ \textbf{0.20\%} & \textbf{14.74\%} $\pm$ \textbf{0.21\%} & \textbf{15.51\%} $\pm$ \textbf{0.18\%} & \textbf{15.92\%} $\pm$ \textbf{0.43\%} \\
\cline{2-9}
        &          & MAE & 20.74 $\pm$ 0.05 & 22.53 $\pm$ 0.10 & \textbf{24.08} $\pm$ \textbf{0.14} & \textbf{26.22} $\pm$ \textbf{0.25} & \textbf{28.65} $\pm$ \textbf{0.23} & \textbf{30.48} $\pm$ \textbf{0.23} \\
PruneGCRN & PeMSD7   & RMSE & 34.74 $\pm$ 0.14 & 38.51 $\pm$ 0.57 & 41.77 $\pm$ 0.56 & \textbf{46.11} $\pm$ \textbf{0.79} & \textbf{50.85} $\pm$ \textbf{0.63} & \textbf{54.41} $\pm$ \textbf{0.59} \\
        &          & MAPE & 8.82\% $\pm$ 0.07\% & 9.58\% $\pm$ 0.08\% & \textbf{10.26\%} $\pm$ \textbf{0.11\%} & \textbf{11.20\%} $\pm$ \textbf{0.14\%} & \textbf{12.35\%} $\pm$ \textbf{0.17\%} & \textbf{13.27\%} $\pm$ \textbf{0.09\%} \\
\cline{2-9}
        &          & MAE & 16.00 $\pm$ 0.11 & 16.71 $\pm$ 0.09 & \textbf{17.41} $\pm$ \textbf{0.17} & \textbf{19.26} $\pm$ \textbf{0.19} & \textbf{21.75} $\pm$ \textbf{0.41} & \textbf{23.68} $\pm$ \textbf{0.73} \\
        & PeMSD8   & RMSE & 25.46 $\pm$ 0.17 & 26.62 $\pm$ 0.13 & \textbf{27.76} $\pm$ \textbf{0.34} & \textbf{30.95} $\pm$ \textbf{0.47} & \textbf{34.84} $\pm$ \textbf{0.51} & \textbf{39.19} $\pm$ \textbf{1.26} \\
        &          & MAPE & 10.33\% $\pm$ 0.08\% & 10.87\% $\pm$ 0.21\% & \textbf{11.14\%} $\pm$ \textbf{0.14\%} & \textbf{12.37\%} $\pm$ \textbf{0.23\%} & \textbf{13.65\%} $\pm$ \textbf{0.36\%} & \textbf{14.29\%} $\pm$ \textbf{0.37\%} \\
\cline{2-9}
        &          & MAE & 1.66 $\pm$ 0.01 & \textbf{1.82} $\pm$ \textbf{0.01} & \textbf{2.00} $\pm$ \textbf{0.01} & \textbf{2.21} $\pm$ \textbf{0.01} & \textbf{2.38} $\pm$ \textbf{0.01} & \textbf{2.46} $\pm$ \textbf{0.01}\\
        & PeMS-Bay & RMSE & 3.82 $\pm$ 0.02 & 4.12 $\pm$ 0.03 & \textbf{4.45} $\pm$ \textbf{0.03} & \textbf{4.82} $\pm$ \textbf{0.03} & \textbf{5.08} $\pm$ \textbf{0.03} & \textbf{5.21} $\pm$ \textbf{0.02}\\
        &          & MAPE & 3.79\% $\pm$ 0.02\% & \textbf{4.14\%} $\pm$ \textbf{0.04\%} & \textbf{4.59\%} $\pm$ \textbf{0.04\%} & \textbf{5.16\%} $\pm$ \textbf{0.04\%} & \textbf{5.54\%} $\pm$ \textbf{0.05\%} & \textbf{5.74\%} $\pm$ \textbf{0.03\%} \\
\hline
\end{tabular}
\caption{Comparison of the results obtained by the model with different pruning methods, datasets and pruning percentages (from 0\% to 95\%). In the first column, Random and Correlation indicate that the mask has been obtained by the respective methods, while PruneGCRN indicates that the mask is learned by the model itself during training.}
\label{tab:comparison_prediction}
\end{table}

The results obtained can be seen in Table \ref{tab:comparison_prediction}. The first column of metrics indicates the result of the model when 0\% of the nodes are removed, so in all cases it is the same mask, and therefore the same metrics. This value has been added to better observe the impact of removing nodes on the prediction. When comparing the results, it can be seen that the random strategy, i.e., using random nodes for the mask, never yields the best results. In contrast, in the case of the correlation strategy, it is the method that obtains the best results for the PeMSD4, PeMSD7, and PeMSD8 datasets, but only when removing 25\% of the nodes. When increasing the number of removed nodes to 50\%, the graph generated during training becomes the best in at least two of the three metrics in all cases. In the case of removing 75\%, 90\%, and 95\% of the nodes, the mask learned during training by PruneGCRN is always the best. Furthermore, if we observe the difference between the mask learned and the other methods, we can see that the greater the number of nodes removed, the greater the improvement provided by the learned mask. It is also important to note that although \textit{correlation} is the best method when removing 25\% of the nodes, from 75\% onwards its error begins to increase significantly, ending up being much worse even than \textit{random}. Thus, it can be seen that the mask generated by the model itself is the best method, improving the existing difference as the complexity of the problem increases by eliminating a greater number of nodes.

Although this work studies the capabilities of obtaining explainability from pruning using PruneGCRN, we also measured the impact that changing the size of the graph has on prediction time and memory usage. To this end, a modified version of the model has been implemented, where the last layer has been changed to predict the removed nodes, also eliminating the bias used to replace the masked value. The results obtained can be seen in Table \ref{tab:comparison_memory_time}. In the case of time, it can be seen how the time is progressively reduced, reaching less than half the time required. In the case of memory used, the difference is even greater, with a reduction of more than 90\% in most datasets. Thus, it can be seen that PruneGCRN bases could be used for graph pruning problems with the aim of reducing computational cost.

\setlength{\tabcolsep}{0.5em}
\begin{table}[H]
\centering
\captionsetup{skip=0.4em}
\def\arraystretch{1.2}%
\begin{tabular}{|c|c|cccccc|}
\hline
Metrics & Dataset & 0\% & 25\% & 50\% & 75\% & 90\% & 95\%\\
\hline
         & PeMSD3   & 25.08 & 15.05 & 13.54 & 11.57 & 10.13 & 9.65\\
         & PeMSD4   & 17.39 & 16.03 & 13.65 & 11.72 & 10.45 & 7.12\\
Time (s) & PeMSD7   & 72.65 & 55.22 & 28.48 & 15.22 & 11.54 & 10.28\\
         & PeMSD8   & 14.94 & 11.20 & 10.56 & 10.37 & 7.12 & 7.23\\
         & PeMS-Bay & 11.58 & 8.63 & 6.68 & 6.07 & 6.03 & 5.95\\
\hline
            & PeMSD3   & 205.48 & 153.59 & 103.28 & 53.27 & 24.75 & 12.38\\
            & PeMSD4   & 178.11 & 133.21 & 92.22 & 49.95 & 20.85 & 13.57\\
Memory (mb) & PeMSD7   & 506.30 & 379.96 & 256.20 & 137.35 & 57.60 & 30.94\\
            & PeMSD8   & 98.07 & 71.36 & 45.96 & 21.66 & 12.02 & 11.99\\
            & PeMS-Bay & 185.62 & 140.77 & 95.50 & 49.40 & 21.57 & 13.79\\
\hline
\end{tabular}
\caption{Comparison of PruneGCRN prediction time and memory based on the dataset and pruning applied.}
\label{tab:comparison_memory_time}
\end{table}

\section{Analysis of results}\label{Analysis}

This section shows the different analyses performed on the model masks. The aim is not only to use the metrics shown in Section \ref{Model_validation} to validate the model, but also to analyze its behavior, showing how the mask learned by the model can be used to obtain explainability. To do this, we first analyzed the distribution of node selection among the different models based on the node selection method. This was done by calculating the frequency with which the different nodes are used by the models in the 10 trainings that were performed for each model on the PEMSD4 dataset.

\begin{figure}[H]
  \centering
  \subfloat[Distribution of node usage frequency when pruned 25\% of them.]
  {
  	\includegraphics[width=5cm]{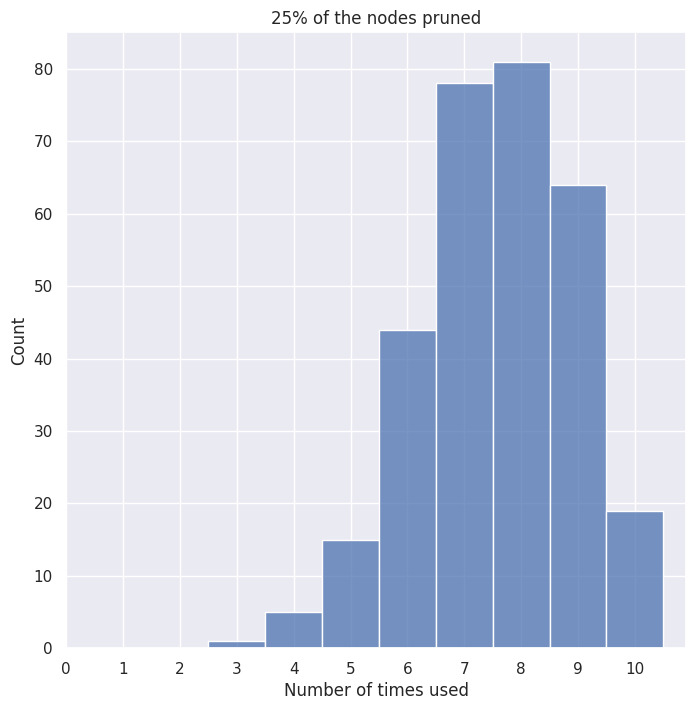}
    \label{fig:random_mask_distribution_25}
  }
  \hfill
  \subfloat[Distribution of node usage frequency when pruned 50\% of them.]
  {
  	\includegraphics[width=5cm]{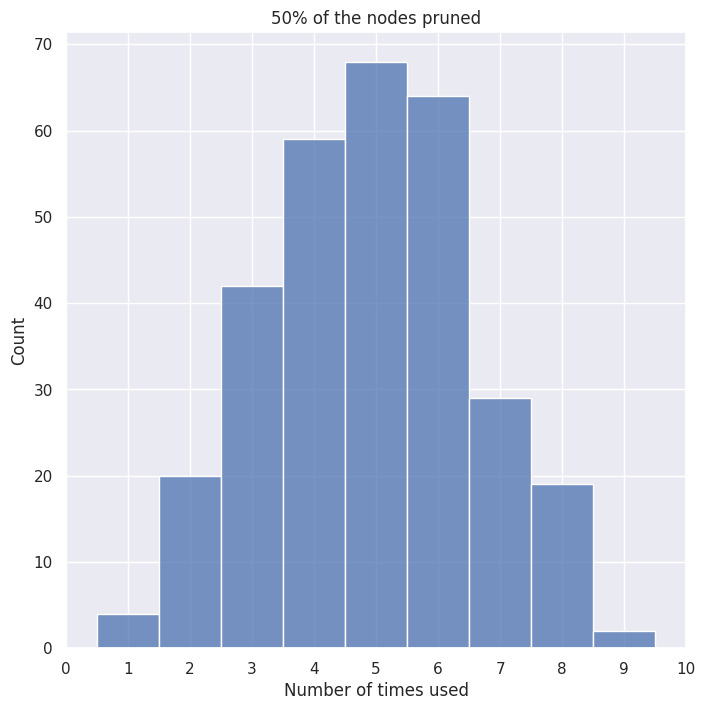}
    \label{fig:random_mask_distribution_50}
  }
  \hfill
  \subfloat[Distribution of node usage frequency when pruned 75\% of them.]
  {
  	\includegraphics[width=5cm]{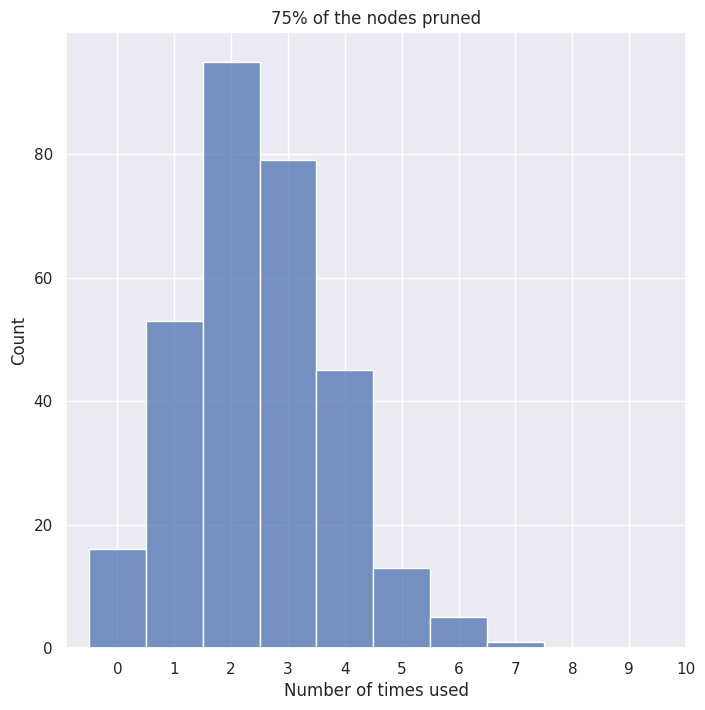}
    \label{fig:random_mask_distribution_75}
  }
  \caption{Distribution of the number of times nodes are used, for different pruning percentages applied. The nodes removed are selected randomly.}
  \label{fig:random_mask_distribution}
\end{figure}

Fig. \ref{fig:random_mask_distribution} shows the distribution of the frequency with which a node is selected when the nodes removed are selected randomly. In  Fig. \ref{fig:random_mask_distribution_25}, we can see that the mean is close to 7.5, which is the expected result, since when 25\% of the nodes are discarded, they have a 75\% probability of being used. An equivalent behavior occurs in Fig. \ref{fig:random_mask_distribution_50} and Fig. \ref{fig:random_mask_distribution_75}.

\begin{figure}[H]
  \centering
  \subfloat[Distribution of node usage frequency when pruned 25\% of them.]
  {
  	\includegraphics[width=5cm]{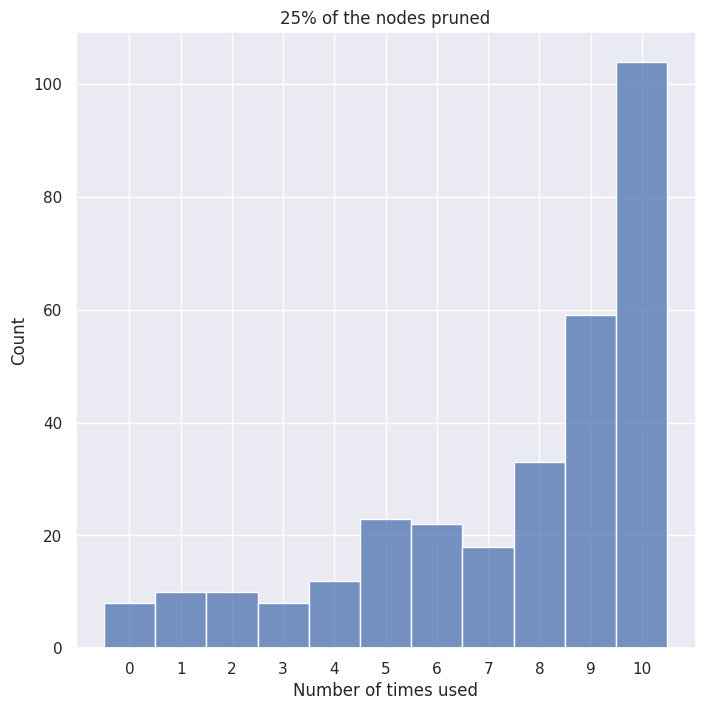}
    \label{fig:training_mask_distribution_25}
  }
  \hfill
  \subfloat[Distribution of node usage frequency when pruned 50\% of them.]
  {
  	\includegraphics[width=5cm]{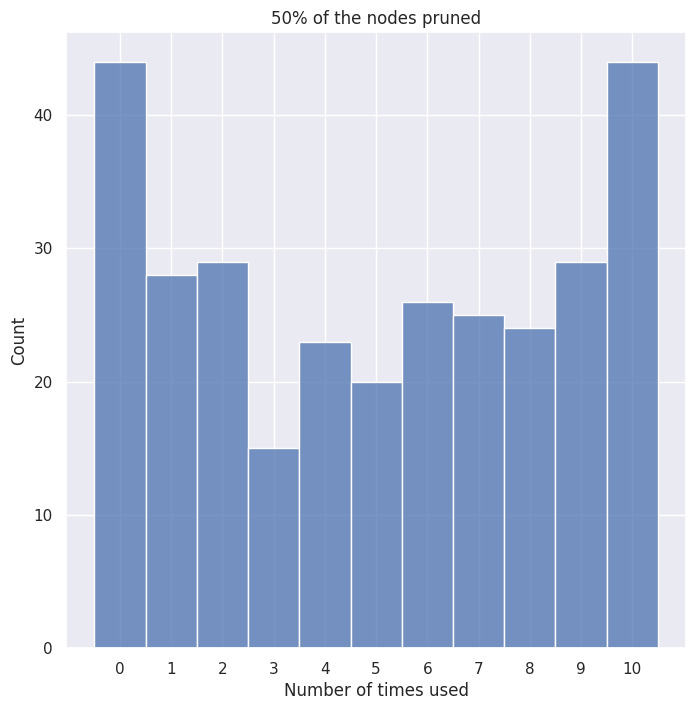}
    \label{fig:training_mask_distribution_50}
  }
  \hfill
  \subfloat[Distribution of node usage frequency when pruned 75\% of them.]
  {
  	\includegraphics[width=5cm]{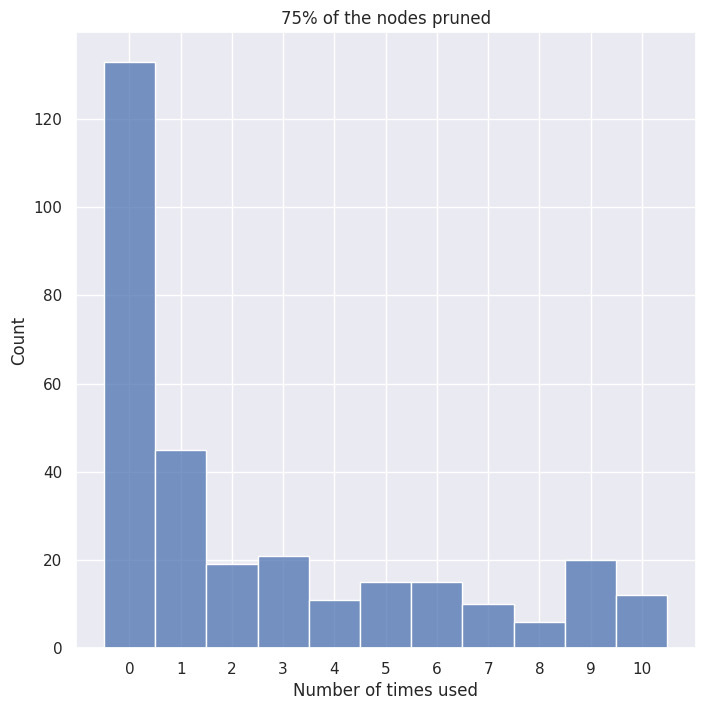}
    \label{fig:training_mask_distribution_75}
  }
  \caption{Distribution of the number of times nodes are used, for different pruning percentages applied. The nodes removed are selected by the model during training.}
  \label{fig:training_mask_distribution}
\end{figure}

These results can be used for visual comparison with Fig. \ref{fig:training_mask_distribution}. This figure shows how frequently a node is used in the mask when the model itself selects the nodes to remove during training. In Fig. \ref{fig:training_mask_distribution_25}, which indicates the use of nodes when 25\% of them are discarded, the most frequent are the nodes that are used in all masks, while there is a queue of rarely used nodes when observing the distribution. In Fig. \ref{fig:training_mask_distribution_50}, where 50\% of the nodes are discarded, it can be seen that the most frequent are the nodes that are always discarded and those that are used in all masks. Finally, in Fig. \ref{fig:training_mask_distribution_75}, where 75\% of the nodes are discarded, the most frequent are the nodes that are discarded in all cases, with a queue that decreases in frequency until reaching the nodes that are used 9 and 10 times, where the quantity rises slightly again. 

\begin{figure}[H]
\centering
\includegraphics[width=11cm]{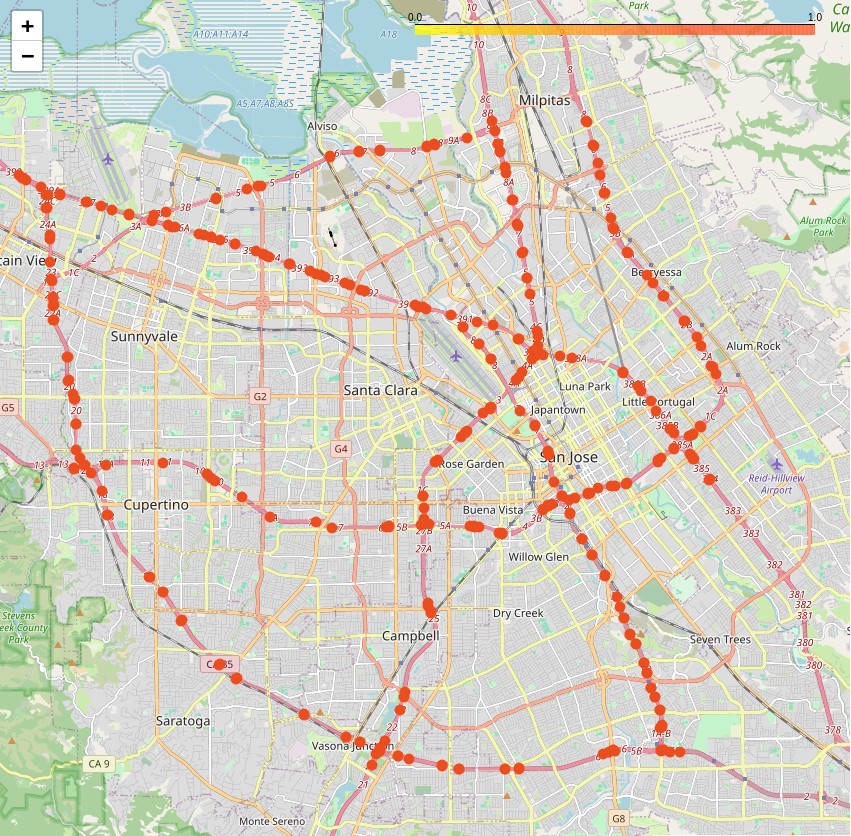} 
\caption{Map showing all the nodes in PeMS-Bay dataset.}
\label{fig:map_all_nodes}
\end{figure}

Thus, Fig. \ref{fig:training_mask_distribution} shows a different behavior from that in Fig. \ref{fig:random_mask_distribution}, indicating that the model appears to have a pattern in node selection and does not perform a random selection. In the case of the mask obtained from the correlation method, this analysis has not been performed, since its distribution does not change, as the correlation between variables is the same across the different trainings.

Although these results show that the selection of mask nodes is not random, and therefore must follow some pattern, a more in-depth analysis of the results is necessary. For this reason, the PeMS-Bay dataset has been used to create various visualizations of the nodes and masks. This dataset was used because it is the only one that includes the coordinates of the radars, which are represented by the nodes of the graphs and enable to perform the analysis.

\begin{figure}[H]
  \centering
  \subfloat[Map showing the nodes in the mask generated selecting random nodes.]
  {
  	\includegraphics[width=7.75cm]{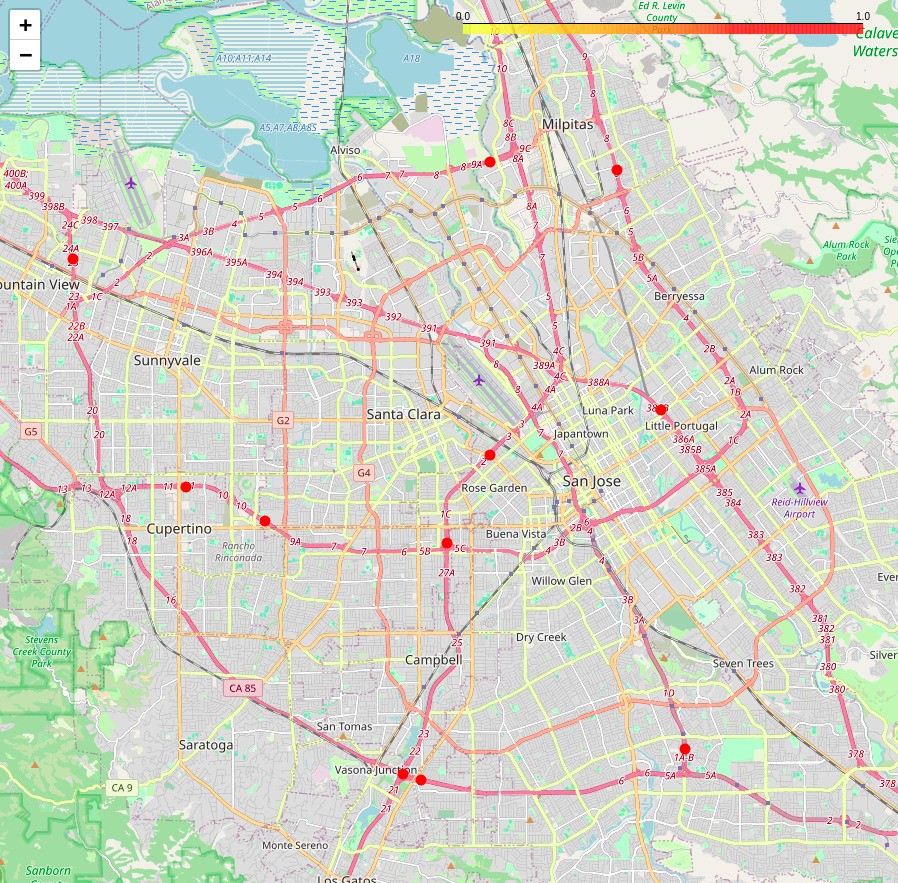}
    \label{fig:map_random_nodes}
  }
  \hfill
  \subfloat[Map showing the nodes in the mask generated during training.]
  {
  	\includegraphics[width=7.75cm]{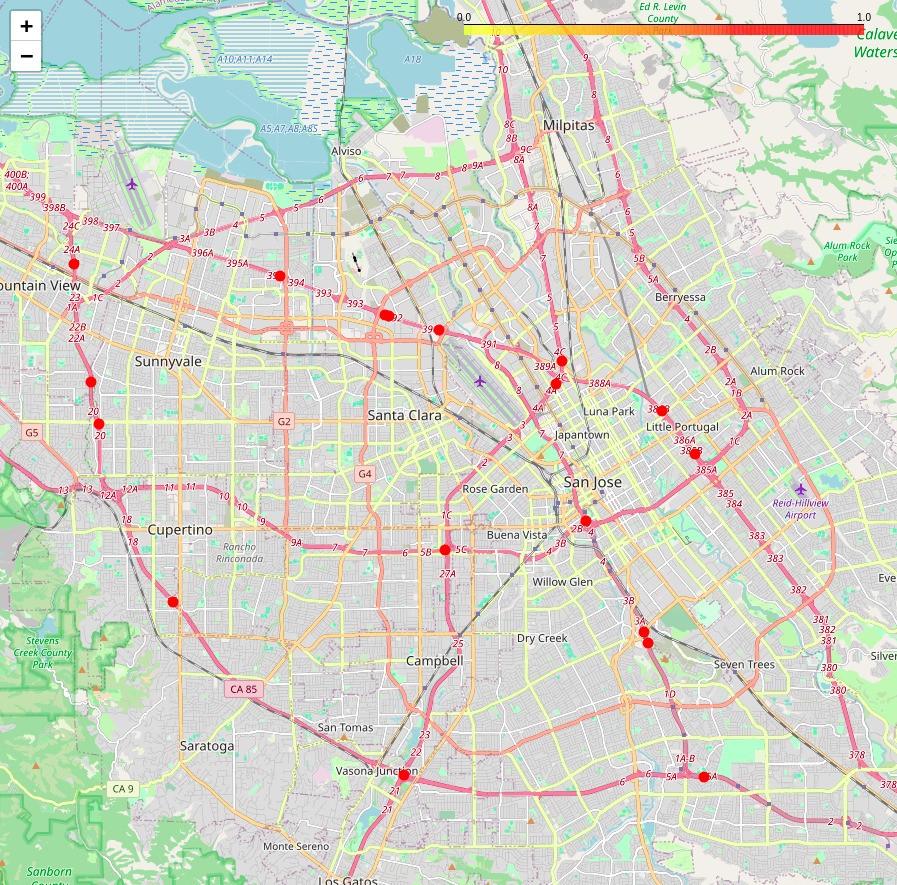}
    \label{fig:map_training_nodes_1}
  }
  \caption{Comparison between the mask generated using random nodes and the mask learned during training.}
  \label{fig:map_random_and_training}
\end{figure}

\begin{figure}[H]
  \centering
  {
  	\includegraphics[width=5.25cm]{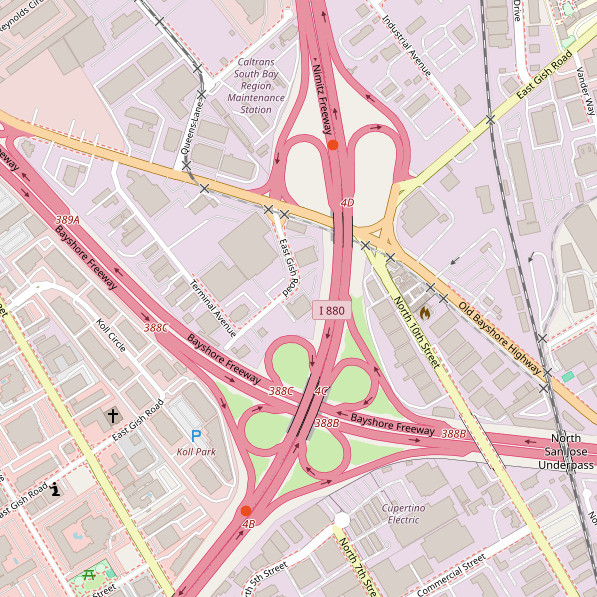}
    \label{fig:map_zoom_1}
  }
  \hfill
  {
  	\includegraphics[width=5.25cm]{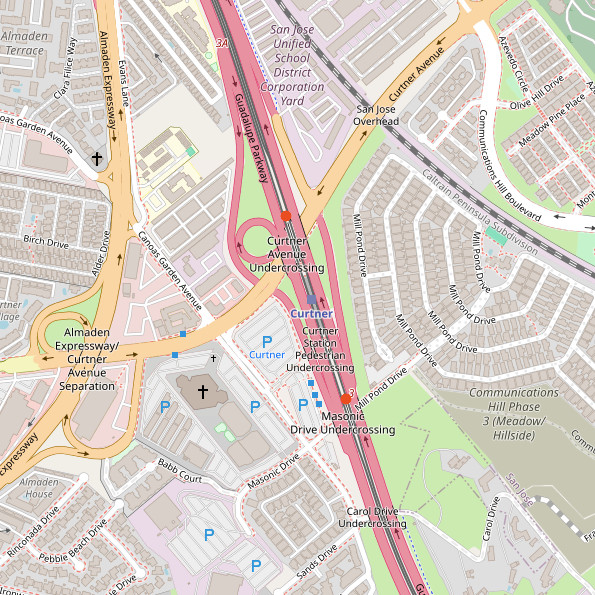}
    \label{fig:map_zoom_2}
  }
  \hfill
  {
  	\includegraphics[width=5.25cm]{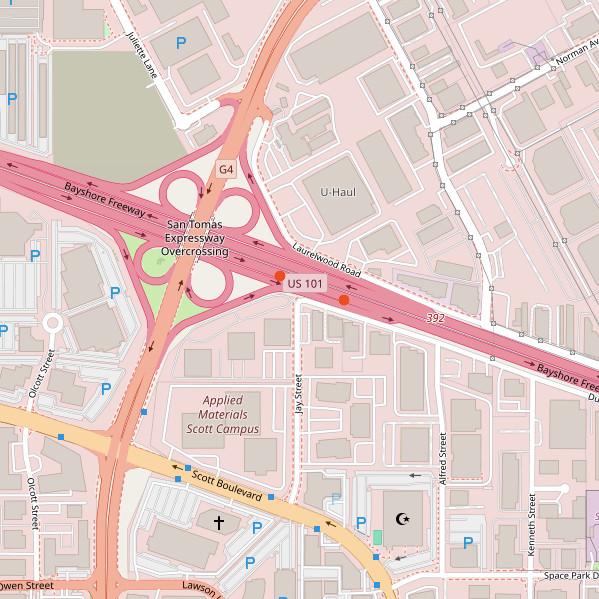}
    \label{fig:map_zoom_3}
  }
  \caption{Comparison between different areas where there are nearby nodes using the mask generated during PruneGCRN training.}
  \label{fig:map_zoom}
\end{figure}

Fig. \ref{fig:map_all_nodes} shows the distribution of all radars on a map of San José. These represent all the nodes in the dataset and all the information available to the model if the mask is not applied. To perform this analysis, the masks generated from the different methods used were explored, using the results obtained when discarding 95\% of the nodes to compare the cases where the difference between methods is most notable.

Fig. \ref{fig:map_random_and_training} shows a comparison between the mask generated selecting random nodes, shown in Fig. \ref{fig:map_random_nodes}, and the mask learned during training, shown in Fig. \ref{fig:map_training_nodes_1}. In the first case, a set of nodes distributed across the map can be seen, with the exception of the two nodes grouped in the south. In contrast, in the second case, there are a greater number of nodes along the highway that crosses the city from east to west. In addition, there are several points where several nodes are located close together. Fig. \ref{fig:map_zoom} shows these cases, where it can be seen that these nodes coincide with areas of the highway that serve as junctions, allowing different traffic sources to be measured. This comparison shows a possible indication that the model is not solely following spatial patterns to select nodes, since in the case of the learned mask, there are areas where nodes have not been used. Furthermore, by selecting only nearby nodes at junctions, the model appears to give importance to these.

\begin{figure}[H]
  \centering
  \subfloat[Map showing the nodes in the mask generated selecting random nodes.]
  {
  	\includegraphics[width=7.75cm]{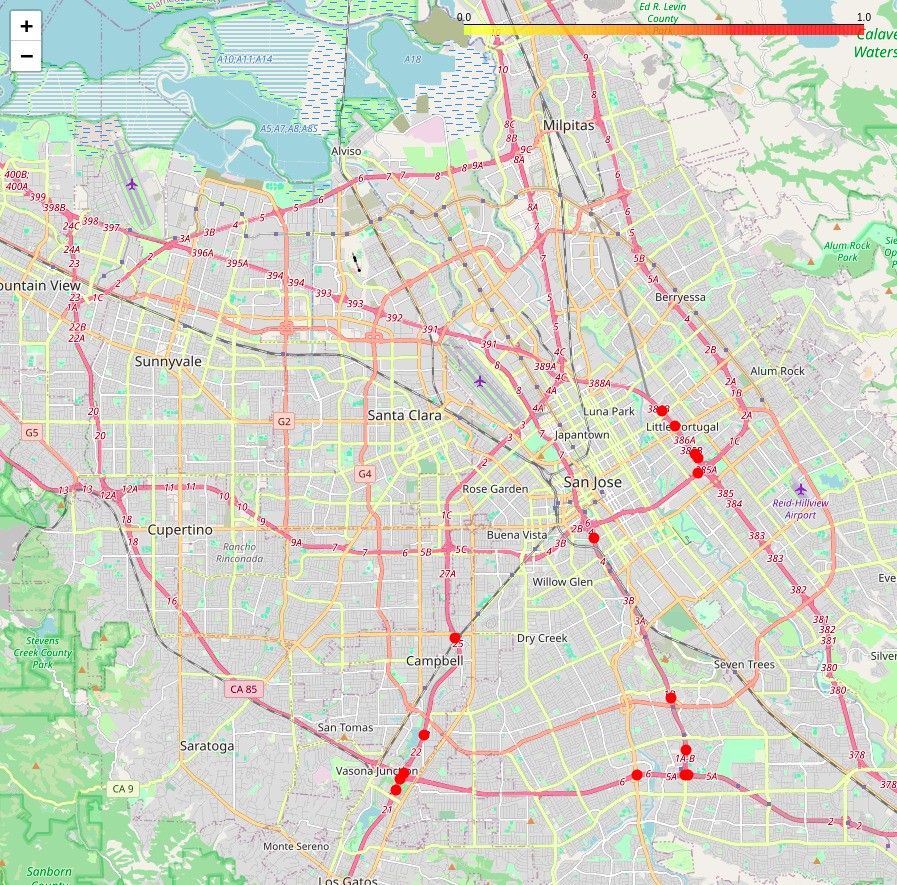}
    \label{fig:map_correlation_nodes}
  }
  \hfill
  \subfloat[Map showing the nodes in the mask generated during training.]
  {
  	\includegraphics[width=7.75cm]{figures/5/map_training_pruning.jpg}
    \label{fig:map_training_nodes_2}
  }
  \caption{Comparison between the mask generated using the correlation between nodes and the mask learned during training.}
  \label{fig:map_correlation_and_training}
\end{figure}

In Fig. \ref{fig:map_correlation_and_training}, we can compare the distribution obtained by selecting nodes based on the correlation between them, Fig. \ref{fig:map_correlation_nodes}, and again, the mask learned during training, Fig. \ref{fig:map_training_nodes_2}. When comparing the two, it can be seen how they differ significantly. In the first case, the distribution is concentrated in a couple of areas located in the south and east of the city. This distribution is possibly one of the reasons why the results shown in Table \ref{tab:comparison_prediction} are much worse when using the correlation method to select nodes with high sparsity values. In this way, the complexity of reducing a graph can be shown visually, since simple methods that might initially seem effective at a theoretical level are confronted with unexpected scenarios due to the complexity of the problem, such as in this case the spatial concentration of the nodes.

\begin{figure}[H]
\centering
\includegraphics[width=14cm]{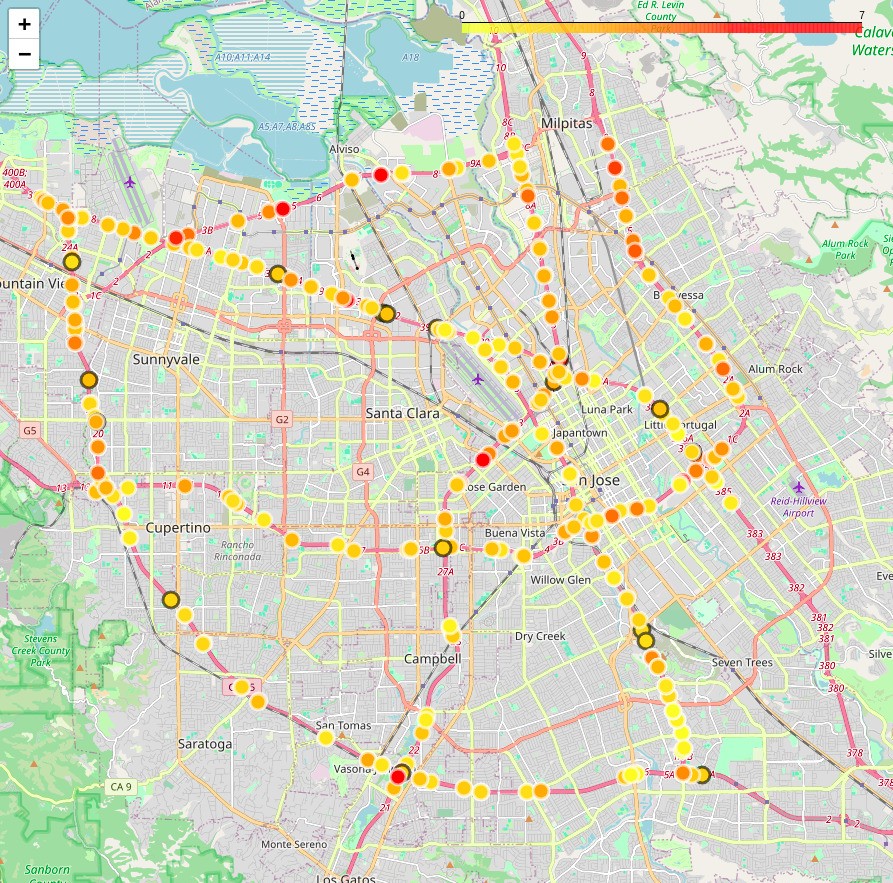} 
\caption{Map showing the discarded nodes (white border) and the selected nodes (black border). The color of the nodes indicates the average error in the prediction, with yellow nodes showing the least error and red nodes indicating the most error.}
\label{fig:map_error_nodes}
\end{figure}

Finally, Fig. \ref{fig:map_error_nodes} shows the nodes used by the model with black borders and the nodes that are discarded with white borders. In addition, the color of the node shows the average prediction error of the model for each node, with yellow indicating a smaller error and red indicating a larger error. In this way, the distribution of the error of the different nodes can be visualized, and it can be checked whether there is a spatial correlation between the distance to the selected nodes and the error in the unselected nodes.

When observing Fig. \ref{fig:map_error_nodes}, no spatial correlation of the prediction error can be seen, but to validate this analysis, Moran's Index has been calculated based on the position of the radars, represented by the nodes, and the error when making the prediction. This analysis seeks to verify, based on a metric, that the model is not only making a spatial selection of nodes, taking those with greater centrality, so that the greater the distance from these, the greater the error. In this way, it is possible to check whether a more complex analysis is being performed and other existing patterns are being searched for.

\setlength{\tabcolsep}{0.5em}
\begin{table}[H]
\centering
\captionsetup{skip=0.4em}
\def\arraystretch{1.2}%
\begin{tabular}{|c|cc|}
\hline
Metrics & Distance & KNN \\
\hline
Moran's Index & 0.0216 & 0.0413 \\
\hline
p-value & 0.1290 & 0.1960 \\
\hline
\end{tabular}
\caption{Comparison of Moran's index and p-value obtained.}
\label{tab:morans_index}
\end{table}

To calculate Moran's Index, it is necessary to generate a graph, which is not present in the dataset, connecting the nodes. Two methods were used to generate this graph: Distance and KNN. Distance uses a minimum distance threshold and connects all nodes below that threshold and KNN connects the two closest neighbors. Once the graphs were generated, Moran's Index was calculated. The results in Table \ref{tab:morans_index} show that all p-values are greater than 0.05, so there is no statistically significant correlation. Therefore, the error is not correlated with the distance to the selected nodes, and the model's selection patterns are different from just spatial distance.

\begin{figure}[H]
\centering
\includegraphics[width=12cm]{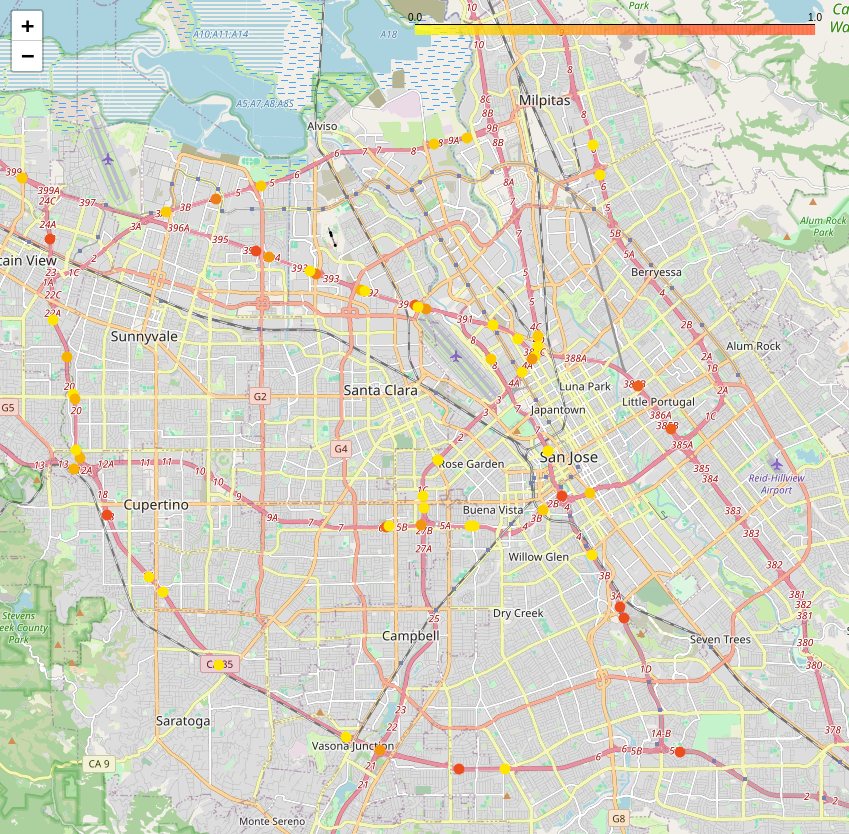} 
\caption{Map showing the frequency with which different nodes are selected by PruneGCRN based on masks obtained in different trainings. The nodes that have been selected least frequently are shown in yellow, and those that have been selected most frequently are shown in red.}
\label{fig:map_10_models}
\end{figure}

In addition, to confirm whether the model follows a consistent pattern between trainings for selecting mask nodes, the masks generated by the model in 10 different trainings were checked. Fig. \ref{fig:map_10_models} shows that although there is variability between trainings, the patterns followed are similar. When analyzing the frequency of use of the different nodes, one node was used by all models, 5 in at least 90\% of cases, and 14 in at least 50\% of cases. By comparing this selection frequency with the 19 nodes that are used per model, out of the 325 existing ones, it can be confirmed that the model follows a pattern for selecting nodes, considering some more relevant than others.

Thus, the analysis carried out has allowed us to explore those elements that the model considers most relevant to the problem.  This has confirmed that node selection is not random and follows complex patterns that are not based solely on spatial distance, but also take into account more complex elements and interdependencies between nodes. In this way, the model is able to reduce the graph, and therefore the components of the problem, in a more complex and effective way than other methods.

\section{Conclusions}\label{Conclusions}

This paper presents PruneGCRN, a model capable of minimizing a spatio-temporal problem to obtain a reduced graph with the most important nodes of the problem. This model seeks to explore the capacity of a deep learning models to provide explainability to problems by finding their key elements and facilitating their analysis by eliminating the most superfluous components. To do this, PruneGCRN is trained with the objective of minimizing prediction error and the number of nodes used at the same time, having to find a balance between both.

For validation purposes, the traffic forecasting problem was used, seeking the most relevant areas for predicting traffic in a city. The results obtained showed that when PruneGCRN learned the mask during training, the prediction were better, in particular when fewer nodes were used. In this way, it was possible to validate the effectiveness of the model in selecting subsets of nodes containing information relevant to the problem. In addition, PruneGCRN has the capacity to open up new lines of research, allowing the design of strategies and models that take advantage of the ability to reduce the size of the graph to generate more compact models, which consume less memory and reduce the cost of computation when making predictions.

\bibliographystyle{unsrtnat}
\bibliography{references}  






\end{document}